\documentclass[10pt,twocolumn,letterpaper]{article}

\usepackage{cvpr}
\usepackage{times}
\usepackage{epsfig}
\usepackage{graphicx}
\usepackage{amsmath}
\usepackage{amssymb}
\usepackage[normalem]{ulem}
\usepackage{afterpage}
\usepackage{booktabs}
\usepackage{color}
\usepackage{verbatim}
\usepackage{soul}
\usepackage{xspace}
\usepackage{comment}
\usepackage{array}
\usepackage{multirow}
\usepackage{url}
\usepackage{epigraph}
\usepackage[toc,page]{appendix}

\cvprfinalcopy

\usepackage[pagebackref=true,breaklinks=true,letterpaper=true,colorlinks,bookmarks=false]{hyperref}

\makeatletter
\DeclareRobustCommand\onedot{\futurelet\@let@token\@onedot}
\def\@onedot{\ifx\@let@token.\else.\null\fi\xspace}

\def\eg{\emph{e.g}\onedot}

\def\iid{\emph{i.i.d}\onedot}
\makeatother
\newcommand{\thickhline}{%
    \noalign {\ifnum 0=`}\fi \hrule height 1pt
    \futurelet \reserved@a \@xhline
}

\newcolumntype{"}{@{\hskip\tabcolsep\vrule width 1.5pt\hskip\tabcolsep}}
\def\cvprPaperID{53} 

\ifcvprfinal\fi

\begin{document}

\title{PixelNet: Representation \textit{of} the pixels, \textit{by} the pixels, and \textit{for} the pixels.} 
\author{Aayush Bansal$^{1}$  \quad     \quad Xinlei Chen$^{1}$ \quad \quad   Bryan Russell$^{2}$  \quad  \quad  Abhinav Gupta$^{1}$  \quad  \quad  Deva Ramanan$^{1}$ \\
$^1$Carnegie Mellon University \quad \quad $^2$Adobe Research \\
{\tt\small{\url{http://www.cs.cmu.edu/~aayushb/pixelNet/}}}
}

\maketitle

\begin{abstract} 
We explore design principles for general pixel-level prediction problems, from low-level edge detection to mid-level surface normal estimation to high-level semantic segmentation. Convolutional predictors, such as the fully-convolutional network (FCN), have achieved remarkable success by exploiting the spatial redundancy of neighboring pixels through convolutional processing. Though computationally efficient, we point out that such approaches are not statistically efficient during learning precisely because spatial redundancy limits the information learned from neighboring pixels. We demonstrate that stratified sampling of pixels allows one to (1) add diversity during batch updates, speeding up learning; (2) explore complex nonlinear predictors, improving accuracy; and (3)  efficiently train state-of-the-art models {\em tabula rasa}  (i.e., ``from scratch'') for diverse pixel-labeling tasks. Our single architecture produces state-of-the-art results for semantic segmentation on PASCAL-Context dataset, surface normal estimation on NYUDv2 depth dataset, and edge detection on BSDS.
\end{abstract}

\section{Introduction}

A number of computer vision problems can be formulated as a dense pixel-wise prediction problem. These include {\bf low-level} tasks such as edge detection~\cite{dollar2013structured,martin2004learning,Xie15} and optical flow~\cite{baker2011database,fischer2015flownet,TeneyH16}, {\bf mid-level} tasks such as depth/normal recovery~\cite{Bansal16,Eigen15,eigen2014depth,saxena20083,Wang15}, and {\bf high-level} tasks such as keypoint prediction~\cite{cao2016realtime,GeorgiaBharathCVPR2014b,Ramanan2007,wei2016cpm}, object detection~\cite{huang2015densebox}, and semantic segmentation~\cite{chen2014semantic,farabet2013learning,Hariharan15,Long15,MostajabiYS15,shotton2006textonboost}. Though such a formulation is attractive because of its generality, one obvious difficulty is the enormous associated output space. For example, a $100\times100$ image with $10$ discrete class labels per pixel yields an output label space of size $10^5$. One strategy is to treat this as a {\em spatially-invariant label prediction} problem, where one predicts a separate label per pixel using a convolutional architecture. Neural networks with convolutional output predictions, also called Fully Convolutional Networks (FCNs)~\cite{chen2014semantic,Long15,matan1991multi,Platt93}, appear to be a promising architecture in this direction.

But is this the ideal formulation of dense pixel-labeling? While {\em computationally efficient} for generating predictions at test time, we argue that it is {\em not statistically efficient} for gradient-based learning. Stochastic gradient descent (SGD) assumes that training data are sampled independently and from an identical distribution (\iid)~\cite{bottou2010large}. Indeed, a commonly-used heuristic to ensure approximately \iid samples is random permutation of the training data, which can significantly improve learnability~\cite{lecun2012efficient}. It is well known that pixels in a given image are highly correlated and not independent~\cite{hyvarinen2009natural}. Following this observation, one might be tempted to randomly permute pixels during learning, but this destroys the spatial regularity that convolutional architectures so cleverly exploit! 
In this paper, we explore the tradeoff between statistical and computational efficiency for convolutional learning, and investigate simply {\em sampling} a modest number of pixels across a small number of images for each SGD batch update, exploiting convolutional processing where possible.

\noindent\textbf{Contributions:} (1) We experimentally validate that, thanks to spatial correlations between pixels, just sampling a small number of pixels per image is sufficient for learning. More importantly, sampling allows us to train end-to-end particular non-linear models not earlier possible, and explore several avenues for improving both the efficiency and performance of FCN-based architectures.  (2) In contrast to the vast majority of models that make use of pre-trained networks, we show that pixel-level optimization can be used to train models {\em tabula rasa}, or ``from scratch'' with simple random Gaussian initialization. Intuitively, pixel-level labels provide a large amount of supervision compared to image-level labels, given proper accounting of correlations. Without using any extra data, our model outperforms previous unsupervised/self-supervised approaches for semantic segmentation on PASCAL VOC-2012~\cite{Everingham10}, and is competitive to fine-tuning from pre-trained models for surface normal estimation.  (3). Using a single architecture and without much modification in parameters, we show state-of-the-art performance for edge detection on BSDS~\cite{amfm_pami2011}, surface normal estimation on NYUDv2 depth dataset~\cite{Silberman12}, and semantic segmentation on the PASCAL-Context dataset~\cite{mottaghi_cvpr14}.

\section{Background}

In this section, we review related work by making use of a unified notation that will be used to describe our architecture. We address the pixel-wise prediction problem where, given an input image $X$, we seek to predict outputs $Y$.  
For pixel location $p$, the output can be binary $Y_p \in \{0,1\}$ (e.g., edge detection), multi-class $Y_p \in \{1,\dots,K\}$ (e.g., semantic segmentation), or real-valued $Y_p \in \mathbb{R}^N$ (e.g., surface normal prediction).  
There is rich prior art in modeling this prediction problem using hand-designed features (representative examples include~\cite{arbelaez2012semantic,carreira2012semantic,dollar2013structured,gould2009decomposing,liu2011nonparametric,munoz2010stacked,russell2009associative,shotton2006textonboost,tighe2010superparsing,tu2010auto,yao2012describing}).

{\bf Convolutional prediction:} We explore {\em spatially-invariant} predictors $f_{\theta,p}(X)$ that are end-to-end trainable over model parameters $\theta$. 
The family of fully-convolutional and skip networks~\cite{matan1991multi,Platt93} are illustrative examples that have been successfully applied to, e.g., edge detection~\cite{Xie15} and semantic segmentation~\cite{byeon2015scene,chen2014semantic,farabet2013learning,fischer2015flownet,Long15,liu2015parsenet,MostajabiYS15,noh2015learning,pinheiro2013recurrent}. Because such architectures still produce separate predictions for each pixel, numerous approaches have explored post-processing steps that enforce spatial consistency across labels via e.g., bilateral smoothing with fully-connected Gaussian CRFs~\cite{chen2014semantic,krahenbuhl2011efficient,zheng2015conditional} or bilateral solvers~\cite{Barron15}, dilated spatial convolutions~\cite{yu2015multi}, LSTMs~\cite{byeon2015scene}, and convolutional pseudo priors~\cite{xie2015convolutional}. In contrast, our work does {\em not} make use of such contextual post-processing, in an effort to see how far a pure ``pixel-level'' architecture can be pushed.

{\bf Multiscale features:} Higher convolutional layers are typically associated with larger receptive fields that capture high-level global context. Because such features may miss low-level details, numerous approaches have built predictors based on multiscale features extracted from multiple layers of a CNN~\cite{denton2015deep,Eigen15,eigen2014depth,farabet2013learning,pinheiro2013recurrent,Wang15}. Hariharan et al.\cite{Hariharan15} use the evocative term ``hypercolumns'' to refer to features extracted from multiple layers that correspond to the same pixel. Let $$h_p(X) = [c_1(p),c_2(p),\ldots, c_M(p)]$$ denote the multi-scale hypercolumn feature computed for pixel $p$, where $c_i(p)$ denotes the feature vector of convolutional responses from layer $i$ centered at pixel $p$ (and where we drop the explicit dependance on $X$ to reduce clutter).
Prior techniques for up-sampling include shift and stitch~\cite{Long15}, converting convolutional filters to dilation operations~\cite{chen2014semantic} (inspired by the {\em algorithme \`{a} trous}~\cite{Holschneider89}), and deconvolution/unpooling~\cite{fischer2015flownet,Long15,noh2015learning}. We similarly make use of multi-scale features, along with {\em sparse} on-demand upsampling of filter responses, with the goal of reducing the memory footprints during learning.

{\bf Pixel-prediction:} One may cast the pixel-wise prediction problem as operating over the hypercolumn features where, for pixel $p$, the final prediction is given by $$f_{\theta,p}(X) = g(h_p(X)).$$ We write $\theta$ to denote both parameters of the hypercolumn features $h$ and the pixel-wise predictor $g$. Training involves back-propagating gradients via SGD to update $\theta$. Prior work has explored different designs for $h$ and $g$. A dominant trend is defining a linear predictor on hypercolumn features, e.g., $g = w \cdot h_p$. FCNs~\cite{Long15} point out that linear prediction can be efficiently implemented in a coarse-to-fine manner by upsampling coarse predictions (with deconvolution) rather than upsampling coarse features. 
DeepLab~\cite{chen2014semantic} incorporates filter dilation and applies similar deconvolution and linear-weighted fusion, in addition to reducing the dimensionality of the fully-connected layers to reduce memory footprint. 
ParseNet~\cite{liu2015parsenet} added spatial context for a layer's responses by average pooling the feature responses, followed by normalization and concatenation. 
HED~\cite{Xie15} output edge predictions from intermediate layers, which are deeply supervised, and fuses the predictions by linear weighting.
Importantly, ~\cite{MostajabiYS15} and \cite{farabet2013learning} are noteable exceptions to the linear trend in that {\em non-linear} predictors $g$ are used. This does pose difficulties during learning -  \cite{MostajabiYS15} precomputes and stores superpixel feature maps due to memory constraints, and so cannot be trained end-to-end. 

{\bf Sampling:} We demonstrate that sparse sampling of hypercolumn features allows for exploration of highly nonlinear $g$, which in turn significantly boosts performance. Our insight is inspired by past approaches that use sampling to training networks for surface normal estimation~\cite{Bansal16} and image colorization \cite{larsson2016learning}, though we focus on general design principles by analyzing the impact of sampling for efficiency, accuracy, and {\em tabula rasa} learning for diverse tasks.

{\bf Accelerating SGD:} There exists a large literature on accelerating stochastic gradient descent. We refer the reader to~\cite{bottou2010large} for an excellent introduction. Though naturally a sequential algorithm that processes one data example at a time, much recent work focuses on mini-batch methods that can exploit parallelism in GPU architectures~\cite{dean2012large} or clusters~\cite{dean2012large}. One general theme is efficient online approximation of second-order methods~\cite{bordes2009sgd}, which can model correlations between input features. Batch normalization~\cite{ioffe2015batch} computes correlation statistics between samples in a batch, producing noticeable improvements in convergence speed. Our work builds similar insights directly into convolutional networks without explicit second-order statistics.

\section{PixelNet}
\label{sec:approach}

This section describes our approach for pixel-wise prediction, making use of the notation introduced in the previous section. We first formalize our pixelwise prediction architecture, and then discuss statistically efficient mini-batch training.

\begin{figure}
\centering
\includegraphics[width=\linewidth]{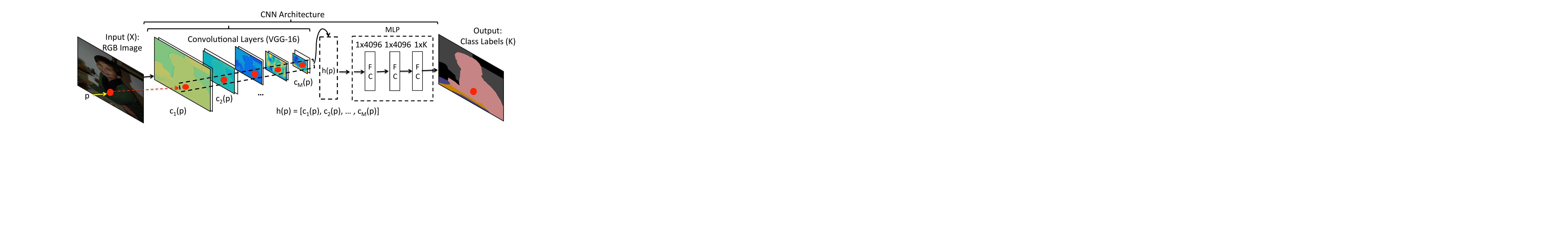}
\caption{\textbf{PixelNet:} We input an image to a convolutional neural network, and extract hypercolumn descriptor for a sampled pixel from multiple convolutional layers. The hypercolumn descriptor is then fed to a multi-layer perceptron (MLP) for the non-linear optimization, and the last layer of MLP outputs the required response for the task. See text for more details about the use of network at training/test time.}
\label{fig:Net}
\end{figure}

{\bf Architecture:} As in past work, our architecture makes use of multiscale convolutional features, which we write as a hypercolumn descriptor: $$h_p= [c_1(p),c_2(p),\ldots, c_M(p) ]$$ We learn a nonlinear predictor $f_{\theta,p} = g(h_p)$ implemented as a multi-layer perceptron (MLP)~\cite{bishop1995neural} defined over hypercolumn features. We use a MLP, which can be implemented as a series of ``fully-connected'' layers followed by ReLU activation functions. Importantly, the last layer must be of size $K$, the number of class labels or real valued outputs being predicted. See Figure~\ref{fig:Net}.

{\bf Sparse predictions:} We now describe an efficient method for generating sparse pixel predictions, which will be used at train-time (for efficient mini-batch generation). Assume that we are given an image $X$ and a sparse set of (sampled) pixel locations $P \subset \Omega$, where $\Omega$ is the set of all pixel positions.
\begin{enumerate}
\item Perform a forward pass to compute dense convolutional responses at all layers $\{c_i(p): \forall i, p \in \Omega \}$
\item For each sampled pixel $p \in P$, compute its hypercolumn feature $h_p$ {\em on demand} as follows: 
\begin{enumerate}
  \item For each layer $i$, compute the 4 discrete locations in the feature map $c_i$ closest to $p$
  \item Compute $c_i(p)$ via bilinear interpolation
\end{enumerate}
\item Rearrange the sparse of hypercolumn features $\{h_p: p \in P\}$ into a matrix for downstream processing (e.g., MLP classification).
\end{enumerate}
The above pipeline only computes $|P|$ hypercolumn features rather than full dense set of size $|\Omega|$. We experimentally demonstrate that this approach offers an excellent tradeoff between amortized computation (to compute $c_i(p)$) and reduced storage (to compute $h_p$). Note that our multi-scale sampling layer simply acts as a selection operation, for which a (sub) gradient can easily be defined. This means that backprop can also take advantage of sparse computations for nonlinear MLP layers and convolutional processing for the lower layers.

{\bf Mini-batch sampling:}
At each iteration of SGD training, the true gradient over the model parameters $\theta$ is approximated by computing the gradient over a relatively small set of samples from the training set. Approaches based on FCN~\cite{Long15} include features for all pixels from an image in a mini-batch.  As nearby pixels in an image are highly correlated~\cite{hyvarinen2009natural}, sampling them will not hurt learning. To ensure a diverse set of pixels (while still enjoying the amortized benefits of convolutional processing), we use a modest number of pixels (${\sim}2,000$) per image, but sample many images per batch. Naive computation of dense grid of hypercolumn descriptors takes almost all of the (GPU) memory, while $2,000$ samples takes a small amount using our sparse sampling layer. This allows us to explore more images per batch, significantly increasing sample diversity. 

{\bf Dense predictions:}  We now describe an efficient method for generating dense pixel predictions with our network, which will be used at test-time. Dense prediction proceeds by following step (1) from above; and instead of sampling in (2) above, we take all the pixels now. This produces a dense grid of hypercolumn features, which are then (3) processed by pixel-wise MLPs implemented as 1x1 filters (representing each fully-connected layer). The memory intensive portion of this computation is the dense grid of hypercolumn features. This memory footprint is reasonable at test time because a single image can be processed at a time, but at train-time, we would like to train on batches containing many images as possible (to ensure diversity).

\section{Analysis}
\label{sec:analysis}

In this section, we analyze the properties of pixel-level optimization using semantic segmentation and surface normal estimation to understand the design choices for pixel-level architectures. We chose the two varied tasks (classification and regression) for analysis to verify the generalizability of these findings. We use a single-scale $224 \times 224$ image as input. 
We also show \textit{sampling} augmented with careful \textit{batch-normalization} can allow for a model to be trained from scratch (without pre-trained ImageNet model as an initialization) for semantic segmentation and surface normal estimation. We explicitly compare the performance of our approach with previous approaches in Section~\ref{sec:exp}. 

\noindent{\bf Default network:} For most experiments we fine-tune a VGG-16 network~\cite{SimonyanZ14a}. VGG-16 has 13 convolutional layers and three fully-connected ({\em fc}) layers. The convolutional layers are denoted as \{$1_1$, $1_2$, $2_1$, $2_2$, $3_1$, $3_2$, $3_3$, $4_1$, $4_2$, $4_3$, $5_1$, $5_2$, $5_3$\}. Following~\cite{Long15}, we transform the last two \emph{fc} layers to convolutional filters\footnote{For alignment purposes, we made a small change by adding a spatial padding of 3 cells for the convolutional counterpart of \emph{fc6} since the kernel size is $7\times 7$.}, and add them to the set of convolutional features that can be aggregated into our multi-scale hypercolumn descriptor. To avoid confusion with the {\bf fc} layers in our MLP, we will henceforth denote the {\em fc} layers of VGG-16 as conv-$6$ and conv-$7$. We use the following network architecture (unless otherwise specified): we extract hypercolumn features from conv-\{$1_2$, $2_2$, $3_3$, $4_3$, $5_3$, $7$\} with on-demand interpolation. We define a MLP over hypercolumn features with 3 fully-connected (fc) layers of size $4,096$ followed by ReLU~\cite{krizhevsky2012imagenet} activations, where the last layer outputs predictions for $K$ classes (with a soft-max/cross-entropy loss) or $K$ outputs with a euclidean loss for regression.

\noindent\textbf{Semantic Segmentation:} We use training images from PASCAL VOC-2012~\cite{Everingham10} for semantic segmentation, and additional labels collected on 8498 images by Hariharan et al.~\cite{hariharan11}. We used the held-out (non-overlapping) validation set to show most analysis. However, at some places we have used the test set where we wanted to show comparison with previous approaches. We report results using the standard metrics of region intersection over union (\textbf{IoU}) averaged over classes (higher is better). We mention it as IoU (V) when using the validation set for evaluation, and IoU (T) when showing on test set.

\noindent\textbf{Surface Normal Estimation:} The NYU Depth v2 dataset~\cite{Silberman12} is used to evaluate the surface normal maps. There are 1449 images, of which 795 are trainval and remaining 654 are used for evaluation. Additionally, there are $220,000$ frames extracted from raw Kinect data. We use the normals of Ladicky et al.\cite{Ladicky14} and Wang et al.\cite{Wang15}, computed from depth data of Kinect, as ground truth for 1449 images and 220K images respectively. We compute six statistics, previously used by \cite{Bansal16,Eigen15,Fouhey13a,fouhey2014unfolding,Fouhey15,Wang15}, over the angular error between the predicted normals and depth-based normals to evaluate the performance -- \textbf{Mean}, \textbf{Median}, \textbf{RMSE}, \textbf{11.25$^\circ$}, \textbf{22.5$^\circ$}, and \textbf{30$^\circ$} --  The first three criteria capture the mean, median, and RMSE of angular error, where lower is better. The last three criteria capture the percentage of pixels within a given angular error, where higher is better. 

\subsection{Sampling}
\label{sec:sampling}

\begin{figure}[t]
\centering
\includegraphics[width=\linewidth]{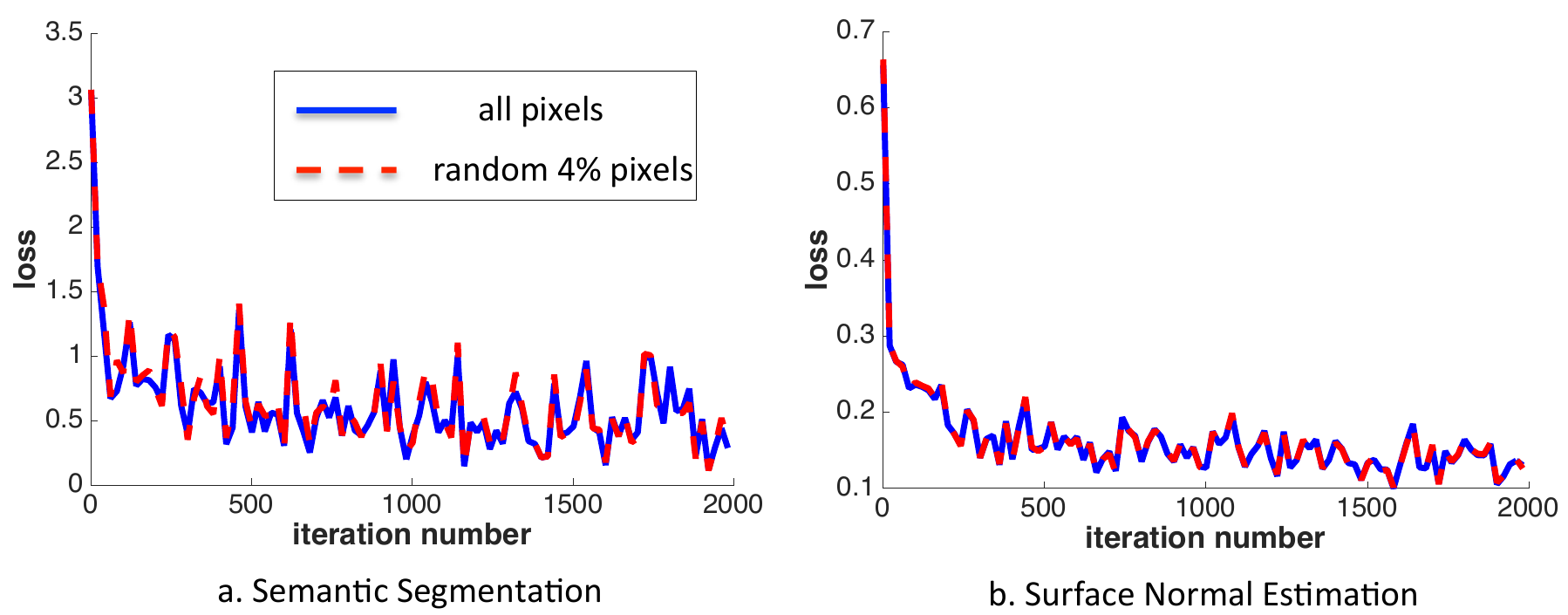}
\caption{Given a fixed number of (5) images per SGD batch, we analyse convergence properties using all pixels vs.\ randomly sampling $4\%$ , or $2,000$ pixels for semantic segmentation and surface normal estimation. This experiment suggest that sampling does not hurt convergence.}
\label{fig:Sampling}
\vspace{-0.5cm}
\end{figure}

We examine how sampling a few pixels from a fixed set of images does not harm convergence. Given a fixed number of (5) images per batch, we find that sampling a small fraction (4\%) of the pixels per image does not affect learnability (Figure ~\ref{fig:Sampling} and Table~\ref{tab:sampling}). This validates our hypothesis that much of the training data for a pixel-level task is correlated within an image, implying that randomly sampling a few pixels is sufficient. Our results are consistent with those reported in Long {\em et al.}~\cite{Long15}, who similarly examine the effect of sampling a fraction (25-50\%) of patches per training image. 

Long {\em et al.}~\cite{Long15} also perform an additional experiment where the total number of pixels in a batch is kept constant when comparing different sampling strategies. While this ensures that each batch will contain more diverse pixels, each batch will also process a larger number of images. If there are no significant computational savings due to sampling, additional images will increase wall-clock time and slow convergence. In the next section, we show that adding additional computation after sampling (by replacing a linear classifier with a multi-layer perceptron) fundamentally changes this tradeoff (Table~\ref{tab:stat_div}).

\begin{table}[t]
\scriptsize{
\setlength{\tabcolsep}{3pt}
\def\arraystretch{1.2}
\center
\begin{tabular}{@{}l |c| c c c c c c }
\toprule
\textbf{Method}  & IoU (V) &  Mean  &   Median & RMSE &  11.25$^\circ$ & 22.5$^\circ$ &  30$^\circ$ \\
\midrule
All Pixels 	     & 44.4 &  25.6  & 19.9   & 32.5  & 29.1  & 54.9  & 66.8  \\
Random 4\% Pixels   & 44.6 &  25.7  & 20.1   & 32.5  & 28.9  & 54.7  & 66.7  \\
\bottomrule
\end{tabular}
\vspace{0.2cm}
\caption{\textbf{Sampling:} We demonstrate that sampling few pixels for each mini-batch yields similar accuracy as using all pixels. The results are computed on models trained for $10$ epochs and $10,000$ iterations for semantic segmentation and surface normal estimation, respectively.}
\label{tab:sampling}
\vspace{-0.2cm}
}
\end{table}

\subsection{Linear vs.\ MLP}

Most previous approaches have focussed on linear predictors combining the information from different convolutional layers (also called `skip-connections'). Here we contrast the performance of non-linear models via MLP with  corresponding linear models. For this analysis, we use a VGG-16 (pre-trained on ImageNet) as initialization and use skip-connections from conv-$\{1_2, 2_2, 3_3, 4_3, 5_3, 7\}$ layers to show the benefits of a non-linear model over a linear model.  We randomly sample $2,000$ pixels per image from a set of five $224{\times}224$ images per SGD iteration for the optimization.

A major challenge in using skip-connections is how to combine the information as the dynamic range varies across the different layers. The top-row in Table~\ref{tab:linear_mlp} shows how the model leads to degenerate outputs for semantic segmentation when `naively' concatenating features from different convolutional layers in a linear model. Similar observation was made by ~\cite{liu2015parsenet}. To counter this issue, previous work has explored normalization~\cite{liu2015parsenet}, scaling~\cite{Hariharan15}, etc.\ We use batch-normalization~\cite{ioffe2015batch} for the convolutional layers before concatenating them to properly train a model with a linear predictor. The middle-row in Table~\ref{tab:linear_mlp} shows how adding batch-normalization allows us to train a linear model for semantic segmentation, and improve the performance for surface normal estimation. While we have to take care of normalization for linear models, we do not need them while using a MLP and can naively concatenate features from different layers. The last row in Table~\ref{tab:linear_mlp} shows the performance on different tasks when using a MLP. Note that performance of linear model (with batch-normalization) is similar to one obtained by Hypercolum~\cite{Hariharan15} (62.7\%), and FCN~\cite{Long15} (62\%).

\begin{table}
\scriptsize{
\setlength{\tabcolsep}{3pt}
\def\arraystretch{1.2}
\center
\begin{tabular}{@{}l |c| c c c c c c }
\toprule
\textbf{Method}  & IoU (T) &  Mean  &   Median & RMSE &  11.25$^\circ$ & 22.5$^\circ$ &  30$^\circ$ \\
\midrule
Linear (no bn) & 3.6 & 24.8   &  19.4  & 31.2 & 28.7  & 56.4 & 68.8  \\
Linear (bn)     & 62.4    &   22.5       & 16.1    &  29.7  &  37.0  &  62.8 &  73.3  \\
MLP 		 & \textbf{67.4}   &   \textbf{19.8}     &	\textbf{12.0}	 &   \textbf{28.0}	 &   \textbf{47.9}     &	   \textbf{70.0}  & \textbf{77.8}		\\ 
\bottomrule
\end{tabular}
\vspace{0.2cm}
\caption{\textbf{Linear vs. MLP:} A multi-layer perceptron over hypercolumn features gives better performance over linear model without requiring normalization/scaling. Note: bn stands for batch-normalization.}
\vspace{-0.2cm}
\label{tab:linear_mlp}
}
\end{table}

\noindent\textbf{\textit{Deconvolution} vs.\ \textit{on-demand} compute: } A naive implementation of our approach is to use deconvolution layers to upsample conv-layers, followed by feature concatenation, and mask out the pixel-level outputs. This is similar to the sampling experiment of Long {\em et al.}~\cite{Long15}. While reasonable for a linear model, naively computing a dense grid of hypercolumn descriptors and processing them with a MLP is impossible if \emph{conv-7} is included in the hypercolumn descriptor (the array dimensions exceed \emph{INT\_MAX}). For practical purposes, if we consider skip-connections only from conv-$\{1_2, 2_2, 3_3, 4_3, 5_3\}$ layers at cost of some performance, naive deconvolution would still take more than \textbf{12X} memory compared to our approach. Slightly better would be masking the dense grid of hypercolumn descriptors before MLP processing, which is still \textbf{8X} more expensive. Most computational savings come from not being required to keep an extra copy of the data required by deconvolution and concatenation operators. Table~\ref{tb:efficiency} highlights the differences in computational requirements between deconvolution vs.\ \textit{on-demand} compute (for the more forgiving setting of \{$1_2$, $3_3$, $5_3$\}-layered hypercolumn features). Clearly, \textit{on-demand} compute requires less resources.

\begin{table}
\setlength{\tabcolsep}{4pt}
\scriptsize{
\begin{center}
\begin{tabular}{ l c c c c c c } 
\toprule
Model & {\#}features & sample (\#) & Memory & Disk Space  & $BPS$ \\
      &              &             &  (MB)  &   (MB)&  \\
\midrule
FCN-32s~\cite{Long15} & 4,096 & 50,176 & 2,010 & 518 & 20.0 \\
FCN-8s~\cite{Long15} & 4,864 &  50,176 & 2,056 & 570 & 19.5 \\
\midrule
\textbf{FCN/Deconvolution} &              &             &    &   &  \\
Linear   & 1,056 &  50,176 & 2,267 & 1,150 & 6.5 \\
MLP & 1,056 &  50,176 & 3,914 & 1,232 & 1.4 \\
\midrule
 \textbf{FCN/Sampling}     &              &             &    &   &  \\
Linear & 1,056 &  2,000 & 2,092 & 1,150 & 5.5 \\
MLP & 1,056 &  2,000 & 2,234 & 1,232 & 5.1 \\
\midrule
\textbf{PixelNet/On-demand}      &              &             &    &   &  \\
Linear & 1,056 &  2,000 & 322 & 60 & 43.3 \\
MLP & 1,056 &  2,000 & 465 & 144 & 24.5 \\
MLP (+\emph{conv-7}) & 5,152 &  2,000 & 1,024 & 686 & 8.8 \\
\bottomrule
\end{tabular}
\vspace{0.2cm}
\caption{\textbf{Computational Requirements:} We record the number of dimensions for hypercolumn features from conv-$\{1_2, 3_3, 5_3\}$,  number of samples (for our model), memory usage, model size on disk, number of mini-batch updates per second ($BPS$ measured by forward/backward passes). We use a single $224\times 224$ image as the input. We compared our network with FCN~\cite{Long15} where a deconvolution layer is used to upsample the result in various settings. Besides FCN-8s and FCN-32s here we first compute the upsampled feature map, and then apply the classifiers for FCN~\cite{Long15}. Clearly from the table, our approach require less computational resources as compared to other settings.}
\label{tb:efficiency}
\vspace{-0.5cm}
\end{center}}
\end{table}

\noindent\textbf{Does statistical diversity matter? } We now analyze the influence of statistical diversity on optimization given a fixed computational budget (7GB memory on a NVIDIA TITAN-X). We train a non-linear model using 1 image $\times$ 40,000 pixels per image  vs.\ 5 images $\times$ 2,000 pixels per image. Table~\ref{tab:stat_div} shows that sampling fewer pixels from more images outperforms more pixels extracted from fewer images. This demonstrates that statistical diversity outweighs the computational savings in convolutional processing when a MLP classifier is used.

\begin{table}
\scriptsize{
\setlength{\tabcolsep}{2pt}
\def\arraystretch{1.2}
\center
\begin{tabular}{@{}l |c c| c c c c c c }
\toprule
\textbf{Method} & IoU$_1$ (V) & IoU$_2$ (V) &  Mean  &   Median & RMSE &  11.25$^\circ$ & 22.5$^\circ$ &  30$^\circ$ \\
\midrule
$1\times40,000$ &   7.9  & 15.5 &  24.8  & 19.5   & 31.6 & 29.7  & 56.1 & 68.5  \\
$5\times2,000$  &   \textbf{38.4}  & \textbf{47.9} &  \textbf{23.4}  & \textbf{17.2}   & \textbf{30.5} & \textbf{33.9}  & \textbf{60.6} & \textbf{71.8}  \\
\bottomrule
\end{tabular}
\vspace{0.2cm}
\caption{\textbf{Statistical Diversity Matters:} For a given computational budget, using diverse set of pixels from more images shows better performance over more pixels from a few images. IoU$_1$ and IoU$_2$  show performance for 10K and 20K iterations of SGD for semantic segmentation. For surface normal estimation, we show performance for 10K iterations of SGD. This suggest that sampling leads to faster convergence.}
\label{tab:stat_div}
\vspace{-0.2cm}
}
\end{table}

\subsection{Training from scratch} Prevailing methods for training deep models make use of a pre-trained (e.g., ImageNet~\cite{Russakovsky15}) model as initialization for fine-tuning for the task at hand. Most network architectures (including ours) improve in performance with pre-trained models. A major concern is the availability of sufficient data to train deep models for pixel-level prediction problems. However, because our optimization is based on randomly-sampled pixels instead of images, there is potentially more unique data available for SGD to learn a model from a random initialization. We show how \textit{sampling} and \textit{batch-normalization} enables models to be trained from scratch. This enables our networks to be used in problems with limited training data, or where natural image data does not apply  (e.g., molecular biology, tissue segmentation etc.\ \cite{nickell2006visual,xu2013automated}). We will show that our results also have implications for unsupervised representation learning~\cite{AgrawalCM15,doersch2015unsupervised,DonahueKD16,GargKC016,Goroshin15,JayaramanG15,Li-ECCV-2016,larsson2016learning,misra2016unsupervised,noroozi2016,Owens0MFT16,pathakCVPR16context,PintoGHPG16,WangG15,zhang2016colorful,zhang2016split}. 

\noindent\textbf{Random Initialization: } We randomly initialize the parameters of a VGG-16 network from a Gaussian distribution. Training a VGG-16 network architecture is not straight forward, and required stage-wise training for the image classification task~\cite{SimonyanZ14a}. It seems daunting to train such a model from scratch for a pixel-level task where we want to learn both coarse and fine information. 
In our experiments, we found batch normalization to be an effective tool for converging a model trained from scratch.

We train the models for semantic segmentation and surface normal estimation. The middle-row in Table ~\ref{tab:rand_init} shows the performance for semantic segmentation and surface normal estimation trained from scratch. The model trained from scratch for surface normal estimation is within ~2-3\% of current state-of-the-art performing method. The model for semantic segmentation achieves 48.7\% on PASCAL VOC-2012 test set when trained from scratch. To the best of our knowledge, these are the best numbers reported on these two tasks when trained from scratch, and exceeds the performance of other unsupervised/self-supervised approaches~\cite{doersch2015unsupervised,larsson2016learning,pathakCVPR16context,WangG15,zhang2016colorful} that required extra ImageNet data~\cite{Russakovsky15}.

\begin{table}
\scriptsize{
\setlength{\tabcolsep}{3pt}
\def\arraystretch{1.2}
\center
\begin{tabular}{@{}l |c| c c c c c c }
\toprule
\textbf{Initialization}  & IoU (T) &  Mean  &   Median & RMSE &  11.25$^\circ$ & 22.5$^\circ$ &  30$^\circ$ \\
\midrule
ImageNet & 67.4 &  19.8     &	12.0	 &   28.2	 &   47.9   &	   70.0  & 77.8	\\
Random  & 48.7 &   21.2 & 13.4   & 29.6 &  44.2 & 66.6  & 75.1  \\
Geometry & 52.4 &  -  &  -  & - & -  & - & -  \\
\bottomrule
\end{tabular}
\vspace{0.2cm}
\caption{\textbf{Initialization:} We study the influence of initialization on accuracy. PixelNet can even be trained reasonably well with random Gaussian initialization or starting from a model trained for surface normal prediction (Geometry).}
\vspace{-0.2cm}
\label{tab:rand_init}
}
\end{table}

\begin{table*}[t!]
\scriptsize{
\setlength{\tabcolsep}{3pt}
\def\arraystretch{1.2}
\center
\begin{tabular}{@{}l c c c c c c c c c c c c c c c c c c c c@{}p{0.3cm}@{}c@{}}
\toprule
\textbf{VOC 2007 test}  & aero  &   bike &  bird & boat &  bottle  &  bus  &  car  &  cat  &  chair & cow & table &  dog  & horse & mbike & person  & plant & sheep & sofa & train & tv &  &\textbf{mAP}\\
\midrule
DPM-v5~\cite{felzenszwalb2010}  & 33.2  &   60.3 & 10.2 & 16.1 &  27.3  &  54.3  &  58.2  &  23.0  &  20.0 & 24.1 & 26.7 &  12.7  & 58.1 & 48.2 & 43.2  & 12.0 & 21.1 & 36.1 & 46.0 & 43.5 &  &33.7\\
HOG+MID ~\cite{BansalSDG15} & 51.7  &  61.5 &  17.9 & 27.0 &  24.0 &  57.5  &  60.2  &  47.9  &  21.1 & 42.2 & 48.9 &  29.8  & 58.3 & 51.9 & 34.3  & 22.2 & 36.8 & 40.2 & 54.3 &  50.9 &  & 41.9\\
\midrule
RCNN-Scratch~\cite{agrawal2014analyzing} & 49.9 & 60.6 & 24.7 & 23.7 & 20.3 & 52.5 & 64.8 & 32.9 & 20.4 & 43.5 & 34.2 & 29.9 & 49.0 & 60.4 & 47.5 & 28.0 & 42.3 & 28.6 & 51.2 & 50.0 &  &40.7\\
VGG-16-Scratch~\cite{doersch2015unsupervised} &
56.1 &        58.6 &        23.3 &        25.7 &        12.8 &        57.8 &        61.2 &        45.2 &        21.4 &        47.1 &        39.5 &        35.6 &        60.1 &        61.4 &        44.9 &        17.3 &        37.7 &        33.2 &        57.9 &        51.2 &       &  42.4 \\
\midrule
VGG-16-Context-v2~\cite{doersch2015unsupervised} & 63.6 &        64.4 &        42.0 &        42.9 &        18.9 &        67.9 &        69.5 &        65.9 &        28.2 &        48.1 &        58.4 &        58.5 &        66.2 &        64.9 &        54.1 &        26.1 &        43.9 &        55.9 &        69.8 &        50.9 &     &   53.0\\
VGG-16-Geometry & 55.9 & 61.6 & 29.5 & 31.1 & 24.0 & 66.3 & 70.6 & 56.7 & 32.4 & 53.2 & 58.5 & 49.4 & 72.1 & 66.4 & 53.6 & 21.8 & 38.6 & 55.1 & 65.4 & 58.2 &  &51.0\\
\midrule
VGG-16-Geom+Seg     & 62.8 & 68.7 & 39.9 & 37.5 & 27.4 & 75.9 & 73.8 & 70.3 & 33.8 & 57.2 & 62.7 & 60.1 & 72.8 & 69.5 & 60.7 & 22.5 & 40.8 & 62.0 & 70.5 & 59.2 &  & 56.4\\
\midrule
VGG-16-ImageNet~\cite{girshick2015fast} & 73.6 & 77.9 & 68.8 & 56.2 & 35.0 & 76.8 & 78.1 & 83.1 & 39.7 & 73.4 & 65.6 & 79.6 & 81.3 & 73.3 & 66.1 & 30.4 & 67.2 & 67.9 & 77.5 & 66.5 &  &66.9\\
\bottomrule
\end{tabular}
\vspace{0.2cm}
\caption{\textbf{Evaluation on VOC-2007:} Our model (VGG-16-Geometry) trained on a few indoor scene examples of NYU-depth dataset performs 9\% better than scratch, and is competitive with ~\cite{doersch2015unsupervised} that used images from ImageNet (without labels) to train. Note that we used 110K iterations of SGD to train our model, and it took less than 3 days. ~\cite{doersch2015unsupervised} required training for around 8 weeks. Finally, we added a minor supervision (VGG-16-Geom+Seg) using the non-overlapping segmentation dataset and improve the performance further by 5\%.}
\label{tab:voc_2007}
}
\vspace{-0.1cm}
\end{table*}

\noindent\textbf{Self-Supervision via Geometry: } We briefly present the performance of models trained from our pixel-level optimization in context of self-supervision. The task of surface normal estimation does not require any human-labels, and is primarily about capturing geometric information. In this section, we explore the applicability of fine-tuning a geometry model (trained from scratch) for more semantic tasks (such as semantic segmentation and object detection).  Table~\ref{tab:rand_init} (last row) and Table~\ref{tab:voc_2007} shows the performance of our approach on semantic segmentation and object detection respectively. Note that the NYU depth dataset is a small indoor scene dataset and does not contain most of the categories present in PASCAL VOC dataset. Despite this, it shows 4\% (segmentation) and 9\% (detection) improvement over naive scratch models. It is best known result for semantic segmentation in an unsupervised/self-supervised manner, and is competitive with the previous unsupervised work~\cite{doersch2015unsupervised} on object detection\footnote{We used a single scale for object detection and use the same parameters as Fast-RCNN except a step-size of 70K, and fine-tuned it for 200K iterations. Doersch et al.\ \cite{doersch2015unsupervised} reports better results in a recent version by the use of multi-scale detection, and smarter initialization and rescaling.} that uses ImageNet (without labels), particularly on indoor scene furniture categories (e.g., chairs, sofa, table, tv, bottle). We posit that geometry is a good cue for unsupervised representation learning as it can learn from a few examples and can even generalize to previously unseen categories. Future work may utilize depth information from videos (c.f.~\cite{Zhang_pami_videodepth}) and use them to train models for surface normal estimation. This can potentially provide knowledge about more general categories. Finally, we add a minor supervision by taking the geometry-based model fine-tuned for segmentation, and further fine-tuning it for object detection. We get an extra 5\% boost over the performance.

\vspace{-0.2cm}
\section{Generalizability\label{sec:exp}}

In this section we demonstrate the generalizability of PixelNet, and apply (with minor modifications) it to the high-level task of semantic segmentation, mid-level surface normal estimation, and the low-level task of edge detection. The in-depth analysis for each of these tasks are in  appendices.

\begin{figure*}
\centering
\includegraphics[width=1\linewidth]{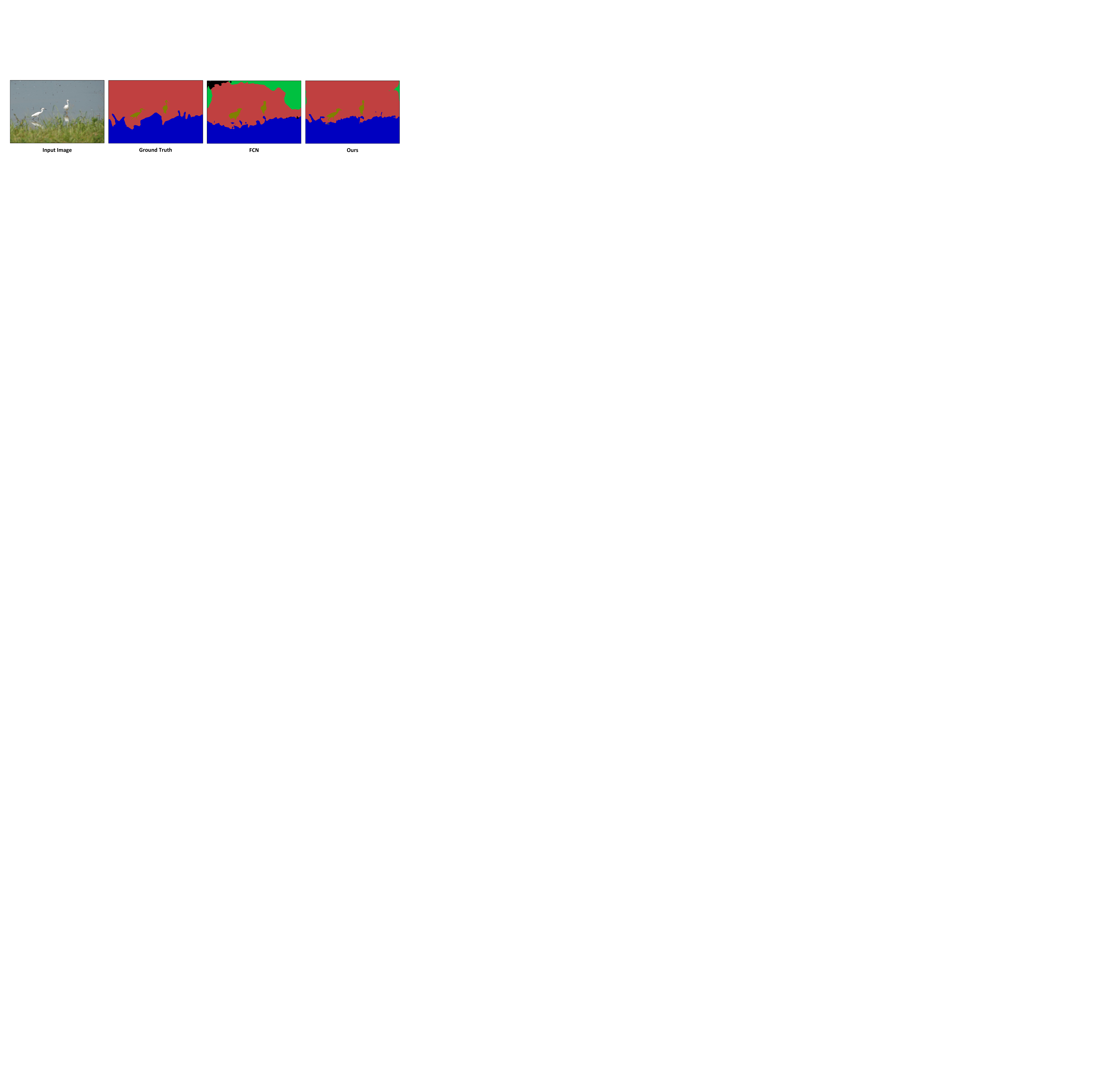}
\caption{Segmentation results on PASCAL-Context 59-class. Our approach uses an MLP to integrate information from both lower (\eg $1_2$) and higher (\eg \emph{conv-7}) layers, which allows us to better capture both global structure (object/scene layout) and fine details (small objects) compared to \emph{FCN-8s}.}
\label{fig:segQ}
\vspace{-0.5cm}
\end{figure*}
 
\subsection{Semantic Segmentation\label{sec:segres}}

\noindent {\bf Training:} For all the experiments we used the publicly available \emph{Caffe} library~\cite{jia2014caffe}. All trained models and code will be released. We make use of ImageNet-pretrained values for all convolutional layers, but train our MLP layers ``from scratch'' with Gaussian initialization ($\sigma=10^{-3}$) and dropout~\cite{srivastava2014dropout} ($r=0.5$). We fix momentum $0.9$ and weight decay $0.0005$ throughout the fine-tuning process. We use the following update schedule (unless otherwise specified): we tune the network for $80$ epochs with  a fixed learning rate ($10^{-3}$), reducing the rate by $10\times$ twice every 8 epochs until we reach $10^{-5}$.

\noindent {\bf Dataset:} The PASCAL-Context dataset~\cite{amfm_pami2011} augments the original sparse set of PASCAL VOC 2010 segmentation annotations~\cite{Everingham10} (defined for 20 categories) to pixel labels for the whole scene. While this requires more than $400$ categories, we followed standard protocol and evaluate on the 59-class and 33-class subsets. The results for PASCAL VOC-2012 dataset~\cite{Everingham10} are in Appendix~\ref{sec:seg}.

\noindent {\bf Evaluation Metrics:} We report results on the standard metrics of pixel accuracy ($AC$) and region intersection over union ($IU$) averaged over classes (higher is better). Both are calculated with DeepLab evaluation tools\footnote{\url{https://bitbucket.org/deeplab/deeplab-public/}}.

\noindent {\bf Results:} Table~\ref{tb:segperf} shows performance of our approach compared to previous work. Our approach without CRF does better than previous approaches based on it. Due to space constraints, we show only one example output in Figure~\ref{fig:segQ} and compare against FCN-8s~\cite{Long15}.  Notice that we capture fine-scale details, such as the leg of birds. More analysis and details are in  Appendix~\ref{sec:seg}. 

\begin{table}
\scriptsize{
\begin{center}
\begin{tabular}{ l  c c  c c } 
\toprule
\multirow{2}{*}{\textbf{Model}} & \multicolumn{2}{c}{\textbf{59-class}} & \multicolumn{2}{c}{\textbf{33-class}} \\
& $AC$ (\%) & $IU$ (\%) & $AC$ (\%) & $IU$ (\%) \\
\midrule
FCN-8s~\cite{DBLP:journals/corr/LongSD14} & 46.5 & 35.1 & 67.6 & 53.5 \\
FCN-8s~\cite{Long15} & 50.7 & 37.8 & - & - \\
DeepLab (v2~\cite{ChenPK0Y16})& - & 37.6 & - & - \\
\midrule
DeepLab (v2) + CRF~\cite{ChenPK0Y16} & - & 39.6 & - & - \\
CRF-RNN~\cite{zheng2015conditional} & - & 39.3 & - & - \\
ConvPP-8~\cite{xie2015convolutional} & - & \textbf{41.0} & - & - \\
\midrule
PixelNet & \textbf{51.5} & \textbf{41.4} & \textbf{69.5} & \textbf{56.9} \\
\bottomrule
\end{tabular}
\vspace{0.2cm}
\caption{\textbf{Evaluation on PASCAL-Context~\cite{amfm_pami2011}:} Most recent approaches~\cite{ChenPK0Y16,xie2015convolutional,zheng2015conditional} except FCN-8s, use spatial context post-processing. We achieve results better than previous approaches without any CRF. CRF post-processing could be applied to any local unary classifier (including our method).}
\label{tb:segperf}
\end{center}}
\vspace{-0.5cm}
\end{table}

\begin{figure*}
\centering
\includegraphics[width=1\linewidth]{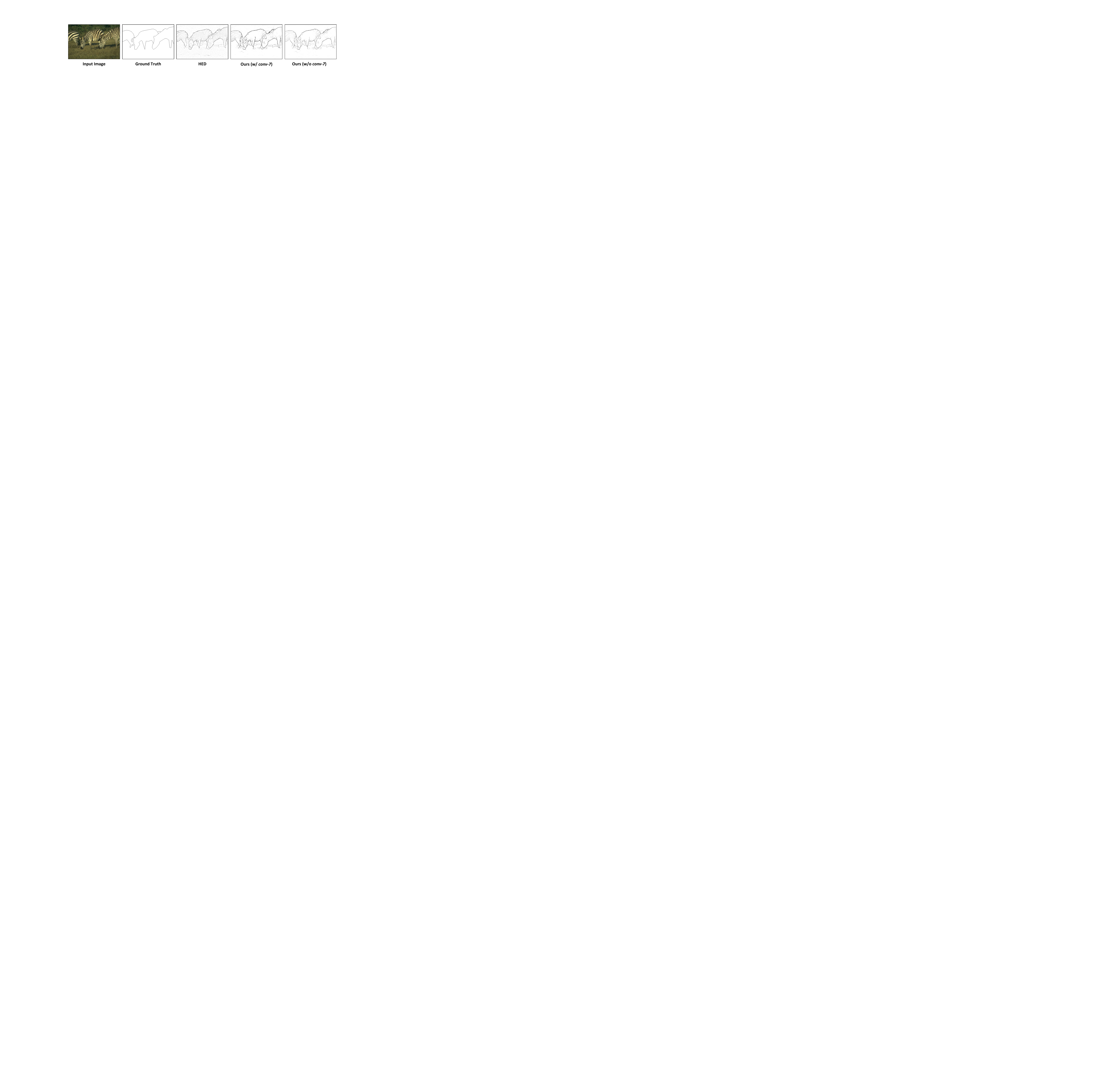}
\caption{Qualitative results for edge detection. Notice that our approach generates more semantic edges for \emph{zebra} compared to HED~\cite{Xie15}. There are more similar examples of \emph{eagle}, and \emph{giraffe} in Appendix~\ref{sec:edges}. Best viewed in the electronic version.}
\label{fig:edges}
\vspace{-0.5cm}
\end{figure*}

\subsection{Surface Normal Estimation\label{sec:surf_normals}} 
 
We use NYU-v2 depth dataset, and same evaluation criteria as defined earlier in Section~\ref{sec:analysis}. We improve the state-of-the-art for surface normal estimation~\cite{Bansal16} using the analysis for general pixel-level optimization. While Bansal et al.\ \cite{Bansal16} extracted hypercolumn features from $1\times1\times4096$ \emph{conv-7} of VGG-16, we provided sufficient padding at \emph{conv-6} to have $4\times4\times4096$ \emph{conv-7}. This provided diversity in \emph{conv-7} features for different pixels in a image instead of same \emph{conv-7} earlier. Further we use a multi-scale prediction to improve the results.

\noindent\textbf{Training: } We use the same network architecture as described earlier. The last \emph{fc}-layer of MLP has ($\sigma=5*10^{-3}$). We set the initial learning rate to $10^{-3}$, reducing the rate by $10\times$ after 50K SGD iteration. The network is trained for 60K iterations.

\noindent\textbf{Results:} Table~\ref{tab:nyud2_scene} compares our improved results with previous state-of-the-art approaches~\cite{Bansal16,Eigen15}. More analysis and details are in Appendix~\ref{sec:normals}.

\begin{table}
\scriptsize{
\setlength{\tabcolsep}{3pt}
\def\arraystretch{1.2}
\center
\begin{tabular}{@{}l c c c c c c }
\toprule
\textbf{NYUDv2 test}  & Mean  &   Median & RMSE &  11.25$^\circ$ & 22.5$^\circ$ &  30$^\circ$ \\
\midrule
Fouhey et al.~\cite{Fouhey13a}	      &  35.3     &	31.2	 &   41.4	 &   16.4      &	  36.6   &	48.2	\\ 
E-F (VGG-16)~\cite{Eigen15} 	      & 20.9      &	13.2	 &    -	 &   44.4     & 67.2	     &	75.9	\\  
UberNet (1-Task)~\cite{Kokkinos16} 	     & 21.4      &	15.6	 &    -	 &   35.3 & 65.9	     &	76.9	\\
MarrRevisited~\cite{Bansal16}		      &	 19.8     &	12.0	 &   28.2	 &   47.9   &	   70.0  & 77.8		\\  
\midrule	
PixelNet		      &	  \textbf{19.2}     &	\textbf{11.3}	 &   \textbf{27.2}	 &   \textbf{49.4}     &	   \textbf{70.7}  & \textbf{78.5}		\\
\bottomrule
\end{tabular}
\vspace{0.2cm}
\caption{\textbf{Evaluation on NYUv2 depth dataset~\cite{Silberman12}:} We improve the previous state-of-the-art~\cite{Bansal16} using the analysis from general pixel-level prediction problems, and multi-scale prediction.}
\label{tab:nyud2_scene}}
\vspace{-0.2cm}
\end{table}

\subsection{Edge Detection\label{sec:edgeres}}
\noindent\textbf{Dataset:} The standard dataset for edge detection is BSDS-500~\cite{amfm_pami2011}, which consists of $200$ training, $100$ validation, and $200$ testing images. Each image is annotated by ${\sim}5$ humans to mark out the contours. We use the same augmented data (rotation, flipping, totaling $9600$ images without resizing) used to train the state-of-the-art Holistically-nested edge detector (HED)~\cite{Xie15}. We report numbers on the testing images. During training, we follow HED and only use positive labels where a consensus ($\ge 3$ out of $5$) of humans agreed.

\begin{table}
\scriptsize{
\begin{center}
\begin{tabular}{ l c c c  }
\toprule
 \  & \textbf{ODS} & \textbf{OIS} & \textbf{AP} \\
\midrule
Human~\cite{amfm_pami2011} & .800 & .800 & - \\
\midrule
SE-Var~\cite{dollar2015fast} & .746 & .767 & .803 \\
OEF~\cite{hallman2015oriented} & .749 & .772 & .817 \\
\midrule
DeepNets~\cite{kivinen2014visual} & .738 & .759 & .758 \\
CSCNN~\cite{hwang2015pixel} & .756 & .775 & .798 \\
HED ~\cite{Xie15} (Updated version) & .788 & .808 & .840 \\
UberNet (1-Task)~\cite{Kokkinos16} & .791 & .809 & \textbf{.849} \\
\midrule
PixelNet (Uniform) & .767 & .786 & .800 \\
PixelNet (Biased)& \textbf{.795} & \textbf{.811} & .830 \\
\bottomrule
\end{tabular}
\vspace{0.2cm}
\caption{\textbf{Evaluation on BSDS~\cite{amfm_pami2011}:} Our approach performs better than previous approaches \textit{specifically} trained for edge detection. Here, we show two results using our architecture: a.\ Uniform; and b.\ Biased. For the former, we randomly sample positive and negative examples while we do a biased sampling of 50\% positives (from a total 10\% positives in edge dataset) and 50\% negatives. As shown, biased sampling improves performance for edge detection. Finally, we achieve F-score of 0.8 which is same as humans.}
\label{tb:bsds}
\end{center}}
\vspace{-0.6cm}
\end{table}

\noindent\textbf{Training:} We use the same baseline network and training strategy defined earlier in Section~\ref{sec:segres}. A sigmoid cross-entropy loss is used to determine the whether a pixel is belonging to an edge or not. Due to the highly skewed class distribution, we also normalized the gradients for positives and negatives in each batch (as in~\cite{Xie15}). In case of skewed class label distribution, sampling offers the flexibility to let the model focus more on the rare classes. 

\noindent{\bf Results:} Table~\ref{tb:bsds} shows the performance of PixelNet for edge detection. The last 2 rows in Table~\ref{tb:bsds} contrast the performance between uniform and biased sampling. Clearly biased sampling toward positives can help the performance. Qualitatively, we find our network tends to have better results for semantic-contours (\eg around an object), particularly after including \emph{conv-7} features. Figure~\ref{fig:edges} shows some qualitative results comparing our network with the HED model. Interestingly, our model explicitly removed the edges inside the \emph{zebra}, but when the model cannot recognize it (\eg its head is out of the picture), it still marks the edges on the black-and-white stripes. Our model appears to be making use of more higher-level information than past work on edge detection. More analysis and details are in Appendix~\ref{sec:edges}.

\section{Discussion} 
We have described a convolutional pixel-level architecture that, with minor modifications, produces state-of-the-art accuracy on diverse high-level, mid-level, and low-level tasks. We demonstrate results on highly-benchmarked semantic segmentation, surface normal estimation, and edge detection datasets. Our results are made possible by careful analysis of computational and statistical considerations associated with convolutional predictors. Convolution exploits spatial redundancy of pixel neighborhoods for efficient computation, but this redundancy also impedes learning. We propose a simple solution based on stratified sampling that injects diversity while taking advantage of amortized convolutional processing. Finally, our efficient learning scheme allow us to explore nonlinear functions of multi-scale features that encode both high-level context and low-level spatial detail, which appears relevant for most pixel prediction tasks.

\begin{appendices}
In Section~\ref{sec:seg} we present extended analysis of PixelNet architecture for semantic segmentation on PASCAL Context~\cite{mottaghi_cvpr14} and PASCAL VOC-2012~\cite{Everingham10}, and ablative analysis for parameter selection. In Section~\ref{sec:normals} we show comparison of improved surface normal with previous state-of-the-art approaches on NYU-v2 depth dataset~\cite{Silberman12}. In Section~\ref{sec:edges} we compare our approach with prior work on edge detection on BSDS~\cite{amfm_pami2011}. Note that we use the default network and training mentioned in the main draft. 

\section{Semantic Segmentation} 
\label{sec:seg}

\noindent {\bf Dataset.} The PASCAL-Context dataset~\cite{amfm_pami2011} augments the original sparse set of PASCAL VOC 2010 segmentation annotations~\cite{Everingham10} (defined for 20 categories) to pixel labels for the whole scene. While this requires more than $400$ categories, we followed standard protocol and evaluate on the 59-class and 33-class subsets. We also evaluated our approach on the standard PASCAL VOC-2012 dataset~\cite{Everingham10} to compare with a wide variety of approaches.

\noindent{\bf Training.} For all the experiments we used the publicly available \emph{Caffe} library~\cite{jia2014caffe}. All trained models and code will be released. 
We make use of ImageNet-pretrained values for all convolutional layers, but train our MLP layers ``from scratch'' with Gaussian initialization ($\sigma=10^{-3}$) and dropout~\cite{srivastava2014dropout}. We fix momentum $0.9$ and weight decay $0.0005$ throughout the fine-tuning process. We use the following update schedule (unless otherwise specified): we tune the network for $80$ epochs with  a fixed learning rate ($10^{-3}$), reducing the rate by $10\times$ twice every 8 epochs until we reach $10^{-5}$. 

\noindent {\bf Qualitative Results.} We show qualitative outputs in Figure~\ref{fig:segQ} and compare against FCN-8s~\cite{Long15}.  Notice that we capture fine-scale details, such as the leg of birds (row 2) and plant leaves (row 3). 

\begin{figure*}
\centering
\includegraphics[width=1\linewidth]{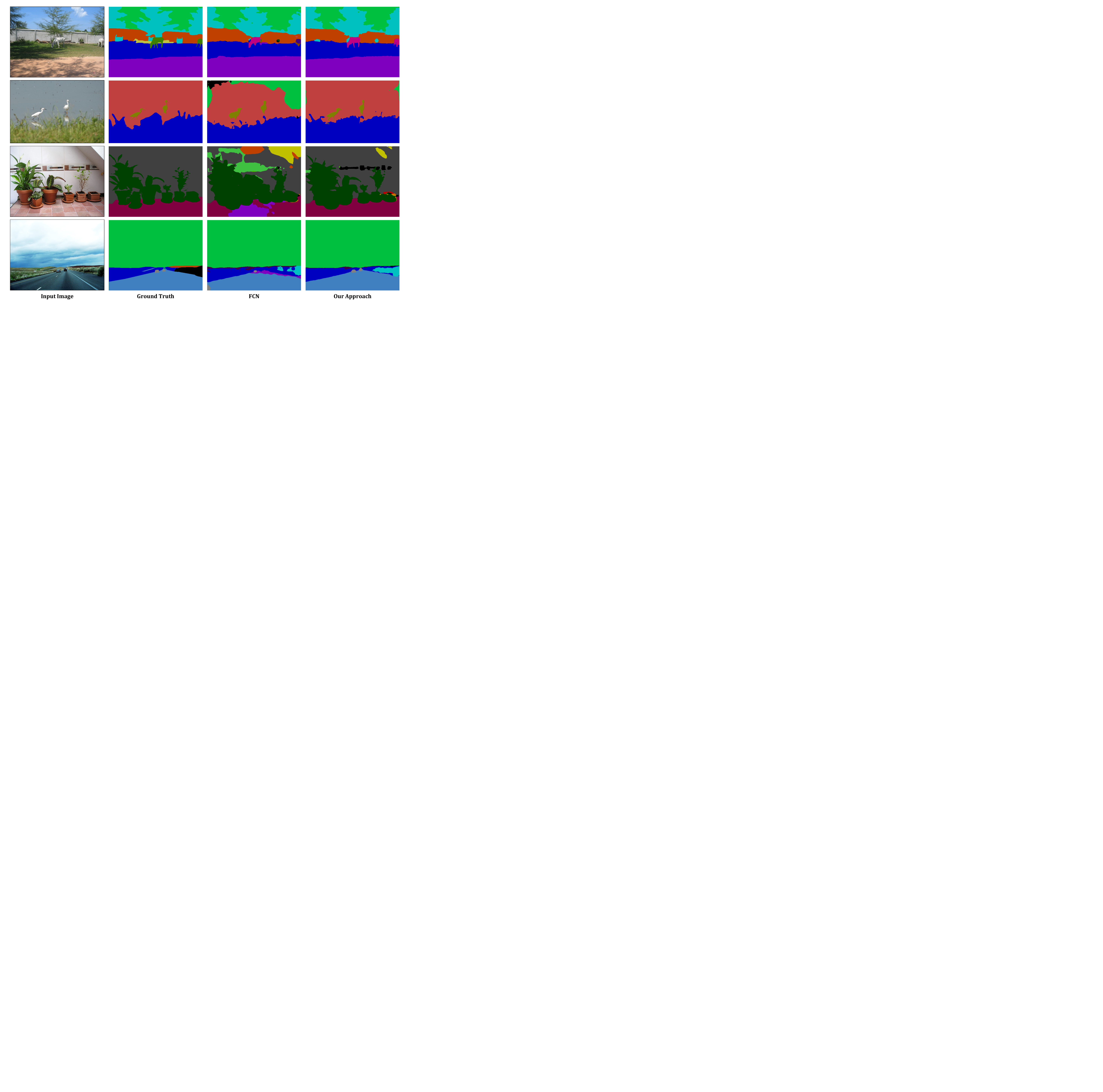}
\caption{Segmentation results on PASCAL-Context 59-class. Our approach uses an MLP to integrate information from both lower (\eg $1_2$) and higher (\eg \emph{conv-7}) layers, which allows us to better capture both global structure (object/scene layout) and fine details (small objects) compared to \emph{FCN-8s}.}
\label{fig:segQ}
\end{figure*}

\noindent {\bf Evaluation Metrics.} We report results on the standard metrics of pixel accuracy ($AC$) and region intersection over union ($IU$) averaged over classes (higher is better). Both are calculated with DeepLab evaluation tools\footnote{\url{https://bitbucket.org/deeplab/deeplab-public/}}.

\noindent {\bf Analysis-1: Dimension of MLP {\em fc} Layers.} We analyze performance as a function of the size of the MLP \emph{fc} layers. We experimented the following dimensions for our {\em fc} layers: $\{1024, 2048, 4096, 6144\}$. Table~\ref{tb:ans1} lists the results. We use 5 images per SGD batch and sample 2000 pixels per image, and conv-$\{1_2, 2_2, 3_3, 4_3, 5_3\}$ for skip connections to do this analysis. We can see that with more dimensions the network tends to learn better, potentially because it can capture more information (and with drop-out alleviating over-fitting~\cite{srivastava2014dropout}). In the main paper we fix the size to $4,096$ as a good trade-off between performance and speed.

\begin{table}[h]
\small{
\centering
\begin{tabular}{ l c c } 
\toprule
\textbf{Dimension} & $AC$ (\%) & $IU$ (\%) \\
\midrule
$1024$ & 41.6 & 33.2 \\
$2048$ & 43.2 & 34.2 \\
$4096$ & 44.0 & 34.9 \\
$6144$ & 44.2 & 35.1 \\
\bottomrule
\end{tabular}
\vspace{0.2cm}
\caption{\textbf{Dimension of MLP fc layers:} We vary the dimension of the MLP \emph{fc} layers on the PASCAL Context 59-class segmentation task from $\{1024,2048,4096,6144\}$. We observe that $4096$ is a good trade-off between performance and speed.}
\label{tb:ans1}}
\end{table}

\noindent {\bf Analysis-2: Number of Mini-batch Samples.} Continuing from our analysis on statistical diversity (Section 4.2 in main paper), we plot performance as a function of the number of sampled pixels per image. In the first sampling experiment, we fix the batch size to $5$ images and sample $\{500, 1000, 2000, 4000\}$ pixels from each image. We use  conv-$\{1_2, 2_2, 3_3, 4_3, 5_3\}$ for skip connections for this analysis. The results are shown in Table~\ref{tb:ans2}. We observe that: 1) even sampling only $500$ pixels per image (on average 2\% of the ${\sim}20,000$ pixels in an image) produces reasonable performance after just $96$ epochs. 2) performance is roughly constant as we increase the number of samples. 

We also perform experiments where the samples are drawn from the same image. When sampling $2000$ pixels from a single image (comparable in size to batch of $500$ pixels sampled from 5 images), performance dramatically drops. This phenomena consistently holds for additional pixels (Table~\ref{tb:ans2}, bottom rows), verifying our central thesis that statistical diversity of samples can trump the computational savings of convolutional processing during learning.

\begin{table}[h]
\small{
\centering
\begin{tabular}{ l  c c } 
\toprule
\textbf{$N{\times}M$} & $AC$ (\%) & $IU$ (\%) \\
\midrule
$500 \times 5$ & 43.7 & 34.8 \\
$1000 \times 5$ & 43.8 & 34.7 \\
$2000 \times 5$ & 43.8 & 34.7 \\
$4000 \times 5$ & 43.9 & 34.9 \\
\midrule
$ 2000 \times 1$ & 32.6 & 24.6 \\
$10000 \times 1$ & 33.3 & 25.2 \\
\bottomrule
\end{tabular}
\vspace{0.2cm}
\caption{\textbf{Varying SGD mini-batch construction: } We vary the SGD mini-batch construction on the PASCAL Context 59-class segmentation task. $N \times M$ refers to a mini-batch constructed from \emph{N} pixels sampled from each of M images (a total of N$\times$M pixels sampled for optimization). We see that a small number of pixels per image (500, or 2\%) are sufficient for learning. Put in another terms, given a fixed budget of N pixels per mini-batch, performance is maximized when spreading them across a large number of images M. This validates our central thesis that statistical diversity trumps the computational savings of convolutional processing during learning.}
\label{tb:ans2}}
\end{table}

\noindent {\bf Analysis-3: Adding conv-$7$.} While our diagnostics reveal the importance of architecture design and sampling, our best results still do not quite reach the state-of-the-art. For example, a single-scale FCN-32s~\cite{Long15}, without any low-level layers, can already achieve $35.1$.  This suggests that their penultimate \emph{conv-7} layer does capture cues relevant for pixel-level prediction. In practice, we find that simply concatenating {\emph{conv-7}} significantly improves performance. 

\begin{table*}
\small{
\begin{center}
\begin{tabular}{ l  c c  c c } 
\toprule
\multirow{2}{*}{\textbf{Model}} & \multicolumn{2}{c}{\textbf{59-class}} & \multicolumn{2}{c}{\textbf{33-class}} \\
& $AC$ (\%) & $IU$ (\%) & $AC$ (\%) & $IU$ (\%) \\
\midrule
FCN-8s~\cite{DBLP:journals/corr/LongSD14} & 46.5 & 35.1 & 67.6 & 53.5 \\
FCN-8s~\cite{Long15} & 50.7 & 37.8 & - & - \\
DeepLab (v2~\cite{ChenPK0Y16})& - & 37.6 & - & - \\
\midrule
DeepLab (v2) + CRF~\cite{ChenPK0Y16} & - & 39.6 & - & - \\
CRF-RNN~\cite{zheng2015conditional} & - & 39.3 & - & - \\
ParseNet~\cite{liu2015parsenet} & - & 40.4 & - & - \\
ConvPP-8~\cite{xie2015convolutional} & - & \textbf{41.0} & - & - \\
\midrule
baseline (conv-\{$1_2$, $2_2$, $3_3$, $4_3$, $5_3$\}) & 44.0 & 34.9 & 62.5 & 51.1 \\
\midrule
conv-\{$1_2$, $2_2$, $3_3$, $4_3$, $5_3$, $7$\} ($0.25$,$0.5$) & 46.7 & 37.1 & 66.6 & 54.8 \\
conv-\{$1_2$, $2_2$, $3_3$, $4_3$, $5_3$, $7$\} ($0.5$) & 47.5 & 37.4 & 66.3 & 54.0 \\
conv-\{$1_2$, $2_2$, $3_3$, $4_3$, $5_3$, $7$\} ($0.5$-$1.0$) & 48.1 & 37.6 & 67.3 & 54.5 \\
conv-\{$1_2$, $2_2$, $3_3$, $4_3$, $5_3$, $7$\} ($0.5$-$0.25$,$0.5$,$1.0$) & \textbf{51.5} & \textbf{41.4} & \textbf{69.5} & \textbf{56.9} \\
\bottomrule
\end{tabular}
\vspace{0.2cm}
\caption{Our final results and baseline comparison on PASCAL-Context. Note that while most recent approaches spatial context post-processing~\cite{ChenPK0Y16,liu2015parsenet,xie2015convolutional,zheng2015conditional}, we focus on the FCN~\cite{Long15} per-pixel predictor as most approaches are its descendants. Also, note that we (without any CRF) achieve results better than previous approaches. CRF post-processing could be applied to any local unary classifier (including our method). Here we wanted to compare with other local models for a ``pure'' analysis.}
\label{tb:segperf}
\end{center}}
\end{table*}

Following the same training process, the results of our model with \emph{conv-7} features are shown in Table~\ref{tb:segperf}. From this we can see that \emph{conv-7} is greatly helping the performance of semantic segmentation. Even with reduced scale, we are able to obtain a similar $IU$ achieved by FCN-8s~\cite{Long15}, without any extra modeling of context~\cite{chen2014semantic,liu2015parsenet,xie2015convolutional,zheng2015conditional}. For fair comparison, we also experimented with single scale training with 1) half scale $0.5\times$, and 2) full scale $1.0\times$ images. We use 5 images per SGD batch, and sample 2000 pixels per image. We find the results are better without $0.25\times$ training, reaching $37.4\%$ and $37.6\%$ $IU$, respectively, even closer to the FCN-8s performance ($37.8\%$ $IU$). For the 33-class setting, we are already doing better with the baseline model plus \emph{conv-7}.

\noindent {\bf Analysis-4: Multi-scale.} All previous experiments process test images at a single scale ($0.25\times$ or $0.5\times$ its original size), whereas most prior work~\cite{chen2014semantic,liu2015parsenet,Long15,zheng2015conditional} use multiple scales from full-resolution images. A smaller scale allows the model to access more context when making a prediction, but this can hurt performance on small objects. Following past work, we explore test-time averaging of predictions across multiple scales. We tested combinations of $0.25\times$, $0.5\times$ and $1\times$. For efficiency, we just fine-tune the model trained on small scales (right before reducing the learning rate for the first time) with an initial learning rate of $10^{-3}$ and step size of $8$ epochs, and end training after $24$ epochs. The results are also reported in Table~\ref{tb:segperf}. Multi-scale prediction generalizes much better ($41.0\%$ $IU$). Note our pixel-wise predictions do not make use of contextual post-processing (even outperforming some methods that post-processes FCNs to do so~\cite{ChenPK0Y16,zheng2015conditional}).

\noindent {\bf Evaluation on PASCAL VOC-2012~\cite{Everingham10}.} We use the same settings, and evaluate our approach on PASCAL VOC-2012. Our approach, without any special consideration of parameters for this dataset, achieves mAP of \textbf{69.7\%}\footnote{Per-class performance is available at \url{http://host.robots.ox.ac.uk:8080/anonymous/PZH9WH.html}.}. This is much better than previous approaches, e.g. 62.7\% for Hypercolumns~\cite{Hariharan15}, 62\% for FCN~\cite{Long15}, ~67\% for DeepLab (without CRF)~\cite{chen2014semantic} etc. Our performance on VOC-2012 is similar to Mostajabi et al~\cite{MostajabiYS15} despite the fact we use information from only 6 layers while they used information from all the layers. In addition, they use a rectangular region of 256$\times$256 (called \textit{sub-scene}) around the super-pixels. We posit that fine-tuning (or back-propagating gradients to conv-layers) enables efficient and better learning with even lesser layers, and without extra \textit{sub-scene} information in an end-to-end framework. Finally, the use of super-pixels in~\cite{MostajabiYS15} inhibit capturing detailed segmentation mask (and rather gives ``blobby'' output), and it is computationally less-tractable to use their approach for per-pixel optimization as information for each pixel would be required to be stored on disk.

\section{Surface Normal Estimation}
\label{sec:normals}

\noindent\textbf{Dataset.} The NYU Depth v2 dataset~\cite{Silberman12} is used to evaluate the surface normal maps. There are 1449 images, of which 795 are trainval and remaining 654 are used for evaluation. Additionally, there are $220,000$ frames extracted from raw Kinect data. We use the normals of Ladicky et al.\cite{Ladicky14} and Wang et al.\cite{Wang15}, computed from depth data of Kinect, as ground truth for 1449 images and 220K images respectively. 

\noindent\textbf{Evaluation Criteria.} We compute six statistics, previously used by \cite{Bansal16,Eigen15,Fouhey13a,fouhey2014unfolding,Fouhey15,Wang15}, over the angular error between the predicted normals and depth-based normals to evaluate the performance -- \textbf{Mean}, \textbf{Median}, \textbf{RMSE}, \textbf{11.25$^\circ$}, \textbf{22.5$^\circ$}, and \textbf{30$^\circ$} --  The first three criteria capture the mean, median, and RMSE of angular error, where lower is better. The last three criteria capture the percentage of pixels within a given angular error, where higher is better. 

\begin{figure*}[t]
\centering
\includegraphics[width=1\linewidth]{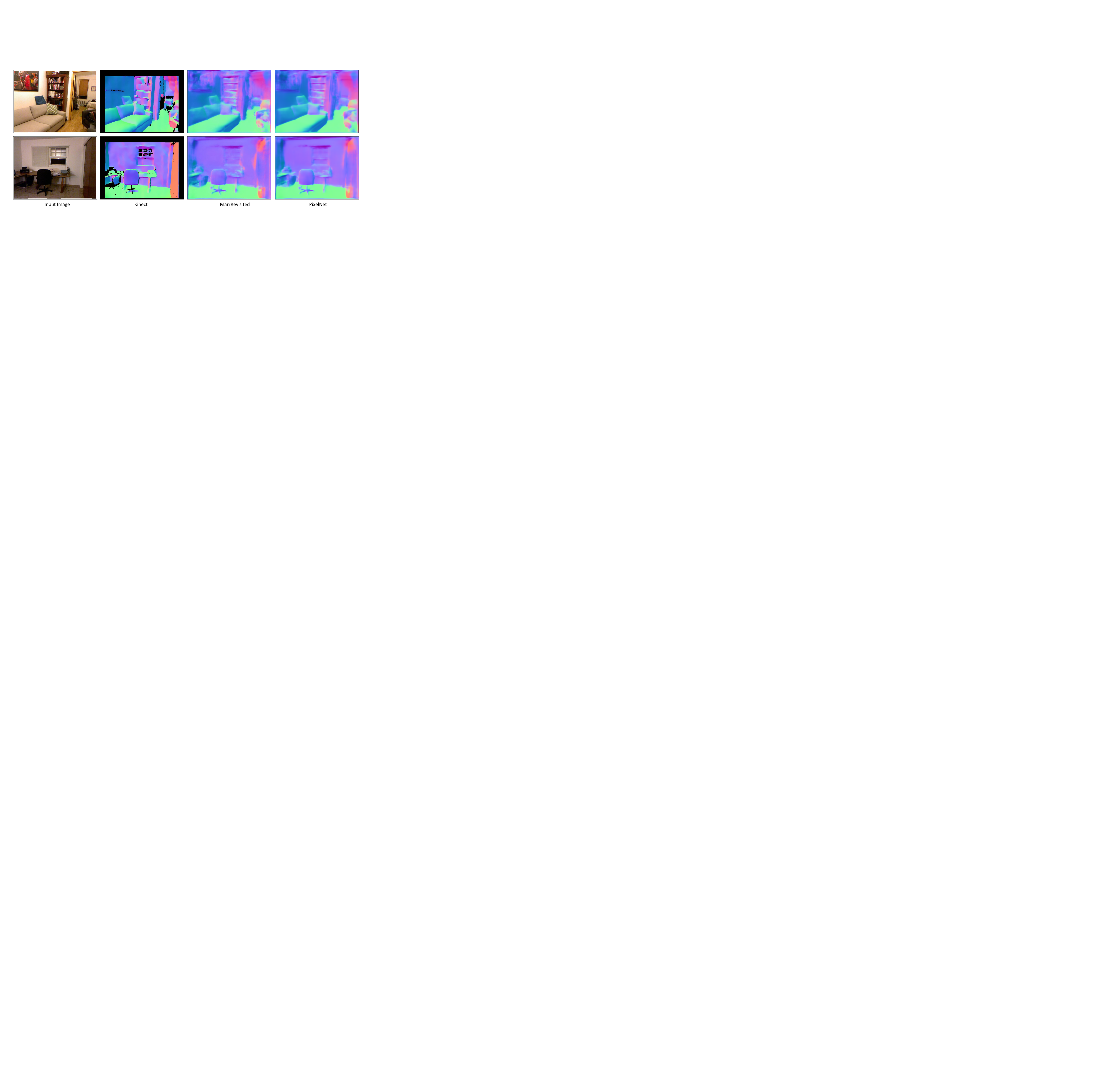}
\caption{Surface normal map estimated on single 2D images from NYU-v2 depth dataset. We show two examples as how the analysis from pixel-level prediction improved the results from MarrRevisited (Bansal et al.\ \cite{Bansal16}). The normals of painting on wall in first row, and room-layout (left side) of second row improved from previous state-of-the-art.}
\label{fig:normals}
\end{figure*}

\noindent\textbf{Qualitative Results.} We show two examples in Figure~\ref{fig:normals} demonstrating where the improvement comes for the surface normal estimation. Note how with multi-scale prediction, the room-layout including painting on the wall improved.

\noindent\textbf{Analysis-1: Global Scene Layout.} We follow Bansal et al.\ \cite{Bansal16}, and present our results both with and without Manhattan-world rectification to fairly compare against previous approaches, as~\cite{Fouhey13a,fouhey2014unfolding,Wang15} use it and~\cite{Eigen15} do not. We rectify our normals using the vanishing point estimates from Hedau et al.~\cite{Hedau2009}. Table~\ref{tab:nyud2_scene} compares our approach with existing work. Similar to Bansal et al.\ \cite{Bansal16}, our approach performs worse with Manhattan-world rectification (unlike Fouhey et al.\ \cite{Fouhey13a}) though it improves slightly on this criteria as well.

\begin{table}
\scriptsize{
\setlength{\tabcolsep}{3pt}
\def\arraystretch{1.2}
\center
\begin{tabular}{@{}l c c c c c c }
\toprule
\textbf{NYUDv2 test}  & Mean  &   Median & RMSE &  11.25$^\circ$ & 22.5$^\circ$ &  30$^\circ$ \\
\midrule
Fouhey et al.~\cite{Fouhey13a}	      &  35.3     &	31.2	 &   41.4	 &   16.4      &	  36.6   &	48.2	\\ 
E-F (VGG-16)~\cite{Eigen15} 	      & 20.9      &	13.2	 &    -	 &   44.4     & 67.2	     &	75.9	\\  
UberNet (1-Task)~\cite{Kokkinos16} 	     & 21.4      &	15.6	 &    -	 &   35.3 & 65.9	     &	76.9	\\
MarrRevisited~\cite{Bansal16}		      &	 19.8     &	12.0	 &   28.2	 &   47.9   &	   70.0  & 77.8		\\  
\midrule	
PixelNet		      &	  \textbf{19.2}     &	\textbf{11.3}	 &   \textbf{27.2}	 &   \textbf{49.4}     &	   \textbf{70.7}  & \textbf{78.5}		\\
\midrule
\midrule
\textbf{Manhattan World} \\ 
Wang et al.~\cite{Wang15}	      &   26.9    &	14.8	 &   - &   42.0     &	 61.2    &	68.2	\\ 
Fouhey et al.~\cite{fouhey2014unfolding}	      & 35.2      &	17.9	 &   49.6	 &    40.5    &	 54.1    &	58.9	\\ 
Fouhey et al.~\cite{Fouhey13a}         &   36.3    &	19.2	 &   50.4	 &  39.2       &	52.9     &	57.8	\\ 
MarrRevisited~\cite{Bansal16}	      &  23.9     &	11.9	 &  35.9	 &  48.4      &	 66.0    &	72.7	\\ 
\midrule
PixelNet		      &	  \textbf{23.6}     &	\textbf{11.8}	 &   \textbf{35.5}	 &   \textbf{48.6}     &	   \textbf{66.3}  & \textbf{73.0}		\\
\bottomrule
\end{tabular}
\vspace{0.2cm}
\caption{\textbf{Evaluation on NYUv2 depth dataset~\cite{Silberman12}: Global Scene Layout.} We improve the previous state-of-the-art~\cite{Bansal16} using the analysis from general pixel-level prediction problems, and multi-scale prediction.}
\label{tab:nyud2_scene}}
\end{table}

\begin{table}
\scriptsize{
\setlength{\tabcolsep}{3pt}
\def\arraystretch{1.2}
\center
\begin{tabular}{@{}l c c  c  c c c }
\toprule
\textbf{NYUDv2 test}  & Mean  &   Median &  RMSE  & 11.25$^\circ$ & 22.5$^\circ$ &  30$^\circ$ \\
\midrule
\textbf{Chair}\\
Wang et al.~\cite{Wang15}  	      &   44.7    &	35.8   &    54.9	 &	 14.2        &	 34.3    &	44.3	\\ 
E-F (AlexNet)~\cite{Eigen15}	      &   38.2    &	32.5   &    46.3         &	  14.4         &	34.9     &	46.6	\\ 
E-F (VGG-16)~\cite{Eigen15}	      &   33.4    &	26.6   &    41.5         &	  18.3         &	43.0     &	55.1	\\
MarrRevisited~\cite{Bansal16}		      &  32.0     &	24.1    &    40.6	 &  21.2	         &	47.3     &	58.5	\\ 
PixelNet & \textbf{31.5}  & \textbf{23.9} &  \textbf{39.9} &   \textbf{21.6} &  \textbf{47.4} &   \textbf{59.1}\\
\midrule
\midrule
\textbf{Sofa}\\
Wang et al.~\cite{Wang15}  	      &  36.0     &	27.6   &   45.4	 &	21.6         &	 42.6    &	53.1	\\ 
E-F (AlexNet)~\cite{Eigen15}	      &   27.0    &	21.3   &	34.0     &	 25.5        &	 52.4    &	63.4	\\ 
E-F (VGG-16)~\cite{Eigen15}	      &   21.6    &	16.8   &	27.3     &	 32.5        &	 63.7    &	76.3	\\ 
MarrRevisited~\cite{Bansal16}		      &   20.9    &	15.9    & 27.0 &	    34.8     &	66.1     &	77.7	\\ 
PixelNet & \textbf{20.2} &  \textbf{15.1} &   \textbf{26.3} &  \textbf{37.2} & \textbf{67.8} &  \textbf{78.9} \\
\midrule
\midrule
\textbf{Bed}\\
Wang et al.~\cite{Wang15}  	      &  28.6     &	18.5   &   38.7	 &	 34.0       &	56.4     &	65.3	\\ 
E-F (AlexNet)~\cite{Eigen15}	      &   23.1    &	16.3   &   30.8	 &	  36.4        &	 62.0    &	72.6	\\ 
E-F (VGG-16)~\cite{Eigen15}	      &   19.9    &	13.6   &   27.1	 &	  43.0        &	 68.2    &	78.3	\\ 
MarrRevisited~\cite{Bansal16}		      &  19.6     &	13.4    &   26.9	 &	43.5         &	69.3     &	79.3	\\ 
PixelNet  & \textbf{19.0} &  \textbf{13.1} &  \textbf{26.1} &   \textbf{44.3} &  \textbf{70.7} & \textbf{80.6} \\
\bottomrule
\end{tabular}
\vspace{3pt}
\caption{\textbf{Evaluation on NYUv2 depth dataset~\cite{Silberman12}: Local Object Layout.} We consistently improve on all metrics for all the objects. This suggest that our approach is able to better capture the fine details present in the scene.
}
\label{tab:nyud2_obj}
}
\end{table}
\noindent\textbf{Analysis-2: Local Object Layout.} Bansal et al.\ \cite{Bansal16} stressed the importance of fine details in the scene generally available around objects. We followed their~\cite{Bansal16} local object evaluation that considers only those pixels which belong to a particular object (such as chair, sofa and bed). Table~\ref{tab:nyud2_obj} shows comparison of our approach with Wang et al.~\cite{Wang15},  Eigen and Fergus~\cite{Eigen15}, and MarrRevisited (Bansal et al.\ \cite{Bansal16}). We consistently improve the performance by \textbf{1-3\%} on all statistics for all the objects.

\section{Edge Detection}
\label{sec:edges}

In this section, we demonstrate that our same architecture can produce state-of-the-art results for low-level edge detection. The standard dataset for edge detection is BSDS-500~\cite{amfm_pami2011}, which consists of $200$ training, $100$ validation, and $200$ testing images. Each image is annotated by ${\sim}5$ humans to mark out the contours. We use the same augmented data (rotation, flipping, totaling $9600$ images without resizing) used to train the state-of-the-art Holistically-nested edge detector (HED)~\cite{Xie15}. We report numbers on the testing images. During training, we follow HED and only use positive labels where a consensus ($\ge 3$ out of $5$) of humans agreed.

\begin{table}
\scriptsize{
\begin{center}
\begin{tabular}{ l c c c  } 
\toprule
 \  & \textbf{ODS} & \textbf{OIS} & \textbf{AP} \\
\midrule
conv-\{$1_2$, $2_2$, $3_3$, $4_3$, $5_3$\} \emph{Uniform} & .767 & .786 & .800 \\
\midrule
conv-\{$1_2$, $2_2$, $3_3$, $4_3$, $5_3$\} (25\%) & .792 & .808 & .826 \\
conv-\{$1_2$, $2_2$, $3_3$, $4_3$, $5_3$\} (50\%) & .791 & .807 & .823 \\
conv-\{$1_2$, $2_2$, $3_3$, $4_3$, $5_3$\} (75\%) & .790 & .805 & .818 \\
\bottomrule
\end{tabular}
\vspace{0.5cm}
\caption{Comparison of different sampling strategies during training. \emph{Top row:} Uniform pixel sampling. \emph{Bottom rows:} Biased sampling of positive examples. We sample a fixed percentage of positive examples ($25\%$,$50\%$ and $75\%$) for each image. Notice a significance difference in performance.}
\label{tb:fgrate}
\end{center}}
\end{table}

\begin{table}
\scriptsize{
\begin{center}
\begin{tabular}{ l c c c  }
\toprule
 \  & \textbf{ODS} & \textbf{OIS} & \textbf{AP} \\
\midrule
Human~\cite{amfm_pami2011} & .800 & .800 & - \\
\midrule
Canny & .600 & .640 & .580 \\
Felz-Hutt~\cite{felzenszwalb2004efficient} & .610 & .640 & .560 \\
\midrule
gPb-owt-ucm~\cite{amfm_pami2011} & .726 & .757 & .696 \\
Sketch Tokens~\cite{lim2013sketch} & .727 & .746 & .780 \\
SCG~\cite{xiaofeng2012discriminatively} & .739 & .758 & .773 \\
\midrule
PMI~\cite{isola2014crisp} & .740 & .770 & .780 \\
\midrule
SE-Var~\cite{dollar2015fast} & .746 & .767 & .803 \\
OEF~\cite{hallman2015oriented} & .749 & .772 & .817 \\
\midrule
DeepNets~\cite{kivinen2014visual} & .738 & .759 & .758 \\
CSCNN~\cite{hwang2015pixel} & .756 & .775 & .798 \\
HED~\cite{Xie15} & .782 & .804 & .833 \\
HED~\cite{Xie15} (Updated version) & .790 & .808 & .811 \\
HED merging~\cite{Xie15} (Updated version) & .788 & .808 & {\bf .840} \\
\midrule
conv-\{$1_2$, $2_2$, $3_3$, $4_3$, $5_3$\} (50\%) & .791 & .807 & .823 \\
conv-\{$1_2$, $2_2$, $3_3$, $4_3$, $5_3$, $7$\} (50\%) & \textbf{.795} & \textbf{.811} & .830 \\
\midrule
conv-\{$1_2$, $2_2$, $3_3$, $4_3$, $5_3$\} (25\%)-($0.5{\times}$,$1.0{\times}$) & .792 & .808 & .826 \\
conv-\{$1_2$, $2_2$, $3_3$, $4_3$, $5_3$, $7$\} (25\%)-($0.5{\times}$,$1.0{\times}$) & \textbf{.795} & \textbf{.811} & .825 \\
\midrule
conv-\{$1_2$, $2_2$, $3_3$, $4_3$, $5_3$\} (50\%)-($0.5{\times}$,$1.0{\times}$) & .791 & .807 & .823 \\
conv-\{$1_2$, $2_2$, $3_3$, $4_3$, $5_3$, $7$\} (50\%)-($0.5{\times}$,$1.0{\times}$) & \textbf{.795} & \textbf{.811} & .830 \\
\midrule
conv-\{$1_2$, $2_2$, $3_3$, $4_3$, $5_3$, $7$\} (25\%)-($1.0{\times}$) & .792 & .808 & .837 \\
conv-\{$1_2$, $2_2$, $3_3$, $4_3$, $5_3$, $7$\} (50\%)-($1.0{\times}$) & .791 & .803 & \textbf{.840} \\
\bottomrule
\end{tabular}
\vspace{0.2cm}
\caption{Evaluation on BSDS~\cite{amfm_pami2011}. Our approach performs better than previous approaches \textit{specifically} trained for edge detection.}
\label{tb:bsds}
\end{center}}
\end{table}

\noindent {\bf Baseline.} We use the same baseline network that was defined for semantic segmentation, only making use of pre-trained \emph{conv} layers. A sigmoid cross-entropy loss is used to determine the whether a pixel is belonging to an edge or not. Due to the highly skewed class distribution, we also normalized the gradients for positives and negatives in each batch (as in~\cite{Xie15}).

\noindent {\bf Training.} We use our previous training strategy, consisting of batches of $5$ images with a total sample size of $10,000$ pixels. Each image is randomly resized to half its scale (so $0.5$ and $1.0$ times) during learning. The initial learning rate is again set to $10^{-3}$. However, since the training data is already augmented, we found the network converges much faster than when training for segmentation. To avoid over-training and over-fitting, we reduce the learning rate at $15$ epochs and $20$ epochs (by a factor of $10$) and end training at $25$ epochs.

\begin{figure}
\centering
\includegraphics[width=\linewidth]{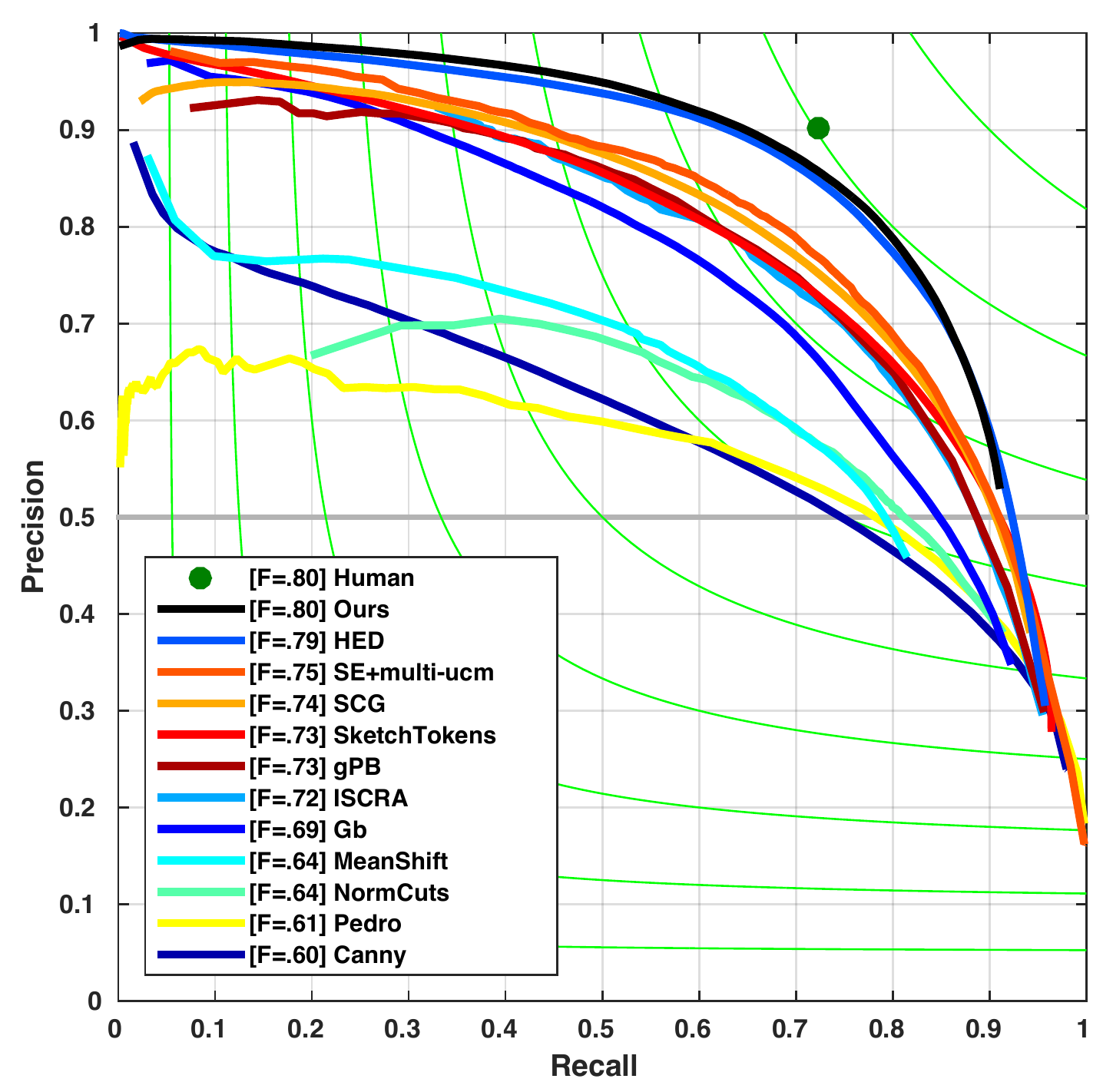}
\caption{Results on BSDS~\cite{amfm_pami2011}. While our curve is mostly overlapping with HED, our detector focuses on more high-level semantic edges. See qualitative results in Fig.\ref{fig:edges}. }
\label{fig:plot}
\end{figure}

\noindent {\bf Baseline Results.} The results on BSDS, along with other concurrent methods, are reported in Table~\ref{tb:bsds}. We apply standard non-maximal suppression and thinning technique using the code provided by~\cite{dollar2013structured}. We evaluate the detection performance using three standard measures: fixed contour threshold (ODS), per-image best threshold (OIS), and average precision (AP).

\begin{figure*}
\centering
\includegraphics[width=1\linewidth]{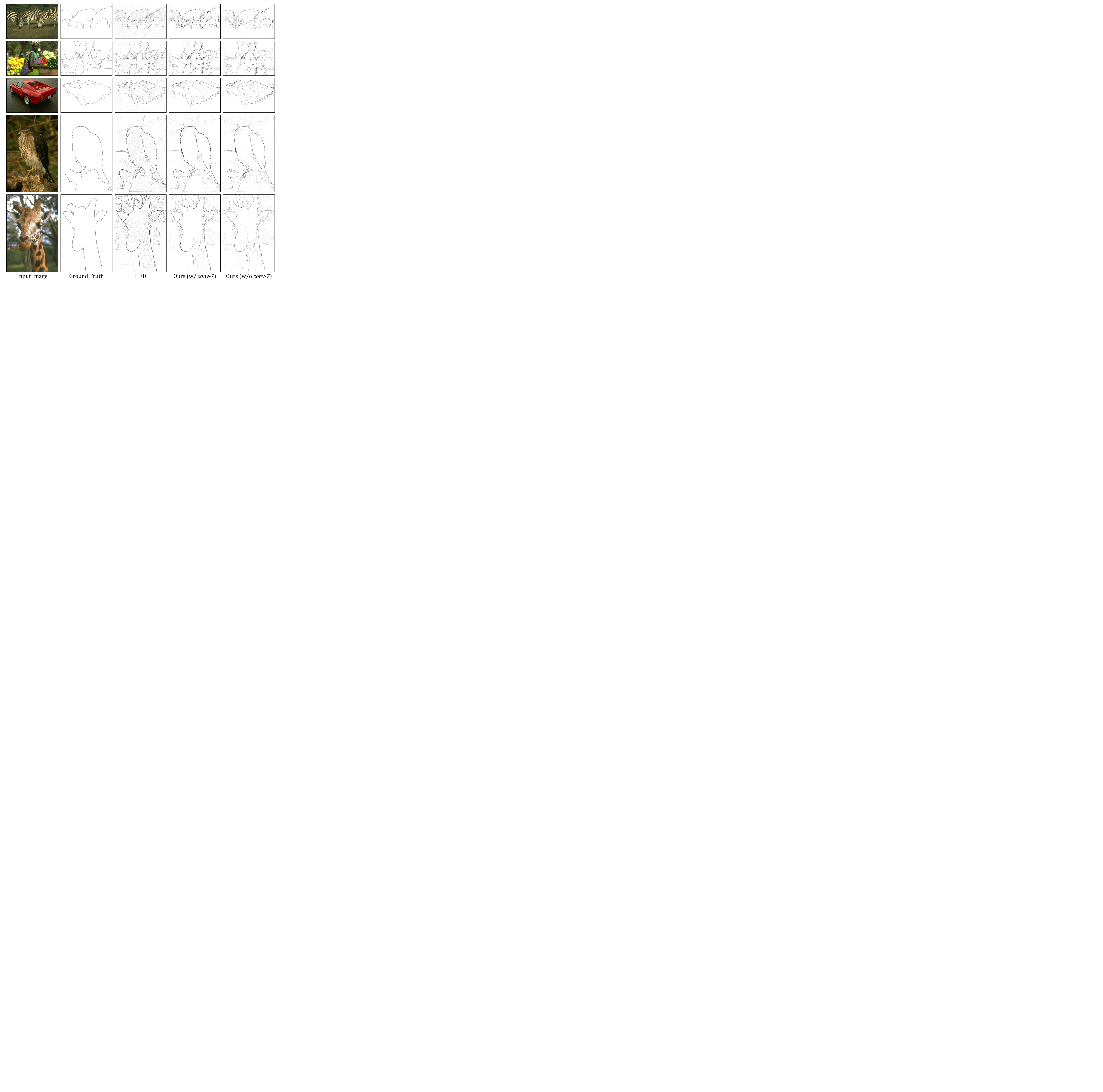}
\caption{Qualitative results for edge detection. Notice that our approach generates more semantic edges for \emph{zebra}, \emph{eagle}, and \emph{giraffe} compared to HED~\cite{Xie15}. Best viewed in the electronic version.}
\label{fig:edges}
\end{figure*}

\noindent {\bf Analysis-1: Sampling.} Whereas uniform sampling sufficed for semantic segmentation~\cite{Long15}, we found the extreme rarity of positive pixels in edge detection required focused sampling of positives. We compare different strategies for sampling a fixed number ($2000$ pixels per image) training examples in Table~\ref{tb:fgrate}. Two obvious approaches are uniform and balanced sampling with an equal ratio of positives and negatives (shown to be useful for object detection~\cite{chen17implementation,girshick2015fast}). We tried ratios of $0.25$, $0.5$ and $0.75$, and found that balancing consistently improved performance\footnote{Note that simple class balancing~\cite{Xie15} in each batch is already used, so the performance gain is \emph{unlikely} from label re-balancing.}. 

\noindent {\bf Analysis-2: Adding \emph{conv-7}.} We previously found that adding features from higher layers is helpful for semantic segmentation. Are such high-level features also helpful for edge detection, generally regarded as a low-level task? To answer this question, we again concatenated \emph{conv-7} features with other \emph{conv} layers \{ $1_2$, $2_2$, $3_3$, $4_3$, $5_3$ \}. Please refer to the results at Table~\ref{tb:bsds}, using the second sampling strategy. We find it still helps performance a bit, but not as significantly for semantic segmentation (clearly a high-level task). Our final results as a single output classifier are very competitive to the state-of-the-art.

Qualitatively, we find our network tends to have better results for semantic-contours (\eg around an object), particularly after including \emph{conv-7} features. Figure~\ref{fig:edges} shows some qualitative results comparing our network with the HED model. Interestingly, our model explicitly removed the edges inside the \emph{zebra}, but when the model cannot recognize it (\eg its head is out of the picture), it still marks the edges on the black-and-white stripes. Our model appears to be making use of much higher-level information than past work on edge detection. 
\end{appendices}

\noindent\textbf{Note to Readers:} An earlier version of this work appeared on arXiv\footnote{\url{https://arxiv.org/pdf/1609.06694.pdf}}. We have incorporated extensive analysis to understand the underlying design principles of general pixel-prediction tasks, and how they can be trained from scratch. We will release the source code and required models on our project page.

\noindent\textbf{Acknowledgements:} This work was in part supported by NSF Grants IIS 0954083, IIS 1618903, and support from Google and Facebook, and Uber Presidential Fellowship to AB.

{\small
\bibliographystyle{ieee}
\bibliography{shortstrings,references}
}

\end{document}